\documentclass[letterpaper]{article} 
\usepackage{aaai23}  
\usepackage{times}  
\usepackage{helvet}  
\usepackage{courier}  
\usepackage[hyphens]{url}  
\usepackage{graphicx} 
\usepackage{natbib}  
\usepackage{caption} 
\usepackage{algorithm}
\usepackage{algorithmic}
\usepackage{newfloat}
\usepackage{listings}
\DeclareCaptionStyle{ruled}{labelfont=normalfont,labelsep=colon,strut=off} 
\floatstyle{ruled}
\newfloat{listing}{tb}{lst}{}
\floatname{listing}{Listing}
\usepackage{subfig}
\usepackage{amssymb}
\usepackage{mathtools}
\usepackage{booktabs}
\usepackage{multirow}
\usepackage{threeparttable}

\usepackage{hyperref}
\hypersetup{
    colorlinks=true,
    citecolor=black,
    linkcolor=black,
    filecolor=black,      
    urlcolor=blue,
}

\title{Diversified and Realistic 3D Augmentation via \\ Iterative Construction, Random Placement, and HPR Occlusion}
\author{
    Jungwook Shin\textsuperscript{\rm 1, 2},
    Jaeill Kim\textsuperscript{\rm 1},
    Kyungeun Lee\textsuperscript{\rm 1},
    Hyunghun Cho\textsuperscript{\rm 1},
    Wonjong Rhee\textsuperscript{\rm 1, 3, 4} 
}

\affiliations{
    \textsuperscript{\rm 1} Department of Intelligence and Information,
    Seoul National University,
    Seoul, 08826, South Korea\\
    \textsuperscript{\rm 2} SK Telecom Co., Ltd, 
    Seoul, 04539, South Korea \\    
    \textsuperscript{\rm 3} Interdisciplinary Program in Artificial Intelligence (IPAI),
    Seoul National University,
    Seoul, 08826, South Korea\\
    \textsuperscript{\rm 4} Artificial Intelligence Institute,
    Seoul National University,
    Seoul, 08826, South Korea\\
    \{jungwook.shin, jaeill0704, ruddms0415, webofthink, wrhee\}@snu.ac.kr
}

\begin{document}

\maketitle

\begin{abstract}
In autonomous driving, data augmentation is commonly used for improving 3D object detection. The most basic methods include insertion of copied objects and rotation and scaling of the entire training frame. Numerous variants have been developed as well. The existing methods, however, are considerably limited when compared to the variety of the real world possibilities. In this work, we develop a diversified and realistic augmentation method that can flexibly construct a whole-body object, freely locate and rotate the object, and apply self-occlusion and external-occlusion accordingly. To improve the diversity of the whole-body object construction, we develop an iterative method that stochastically combines multiple objects observed from the real world into a single object. Unlike the existing augmentation methods, the constructed objects can be randomly located and rotated in the training frame because proper occlusions can be reflected to the whole-body objects in the final step. Finally, proper self-occlusion at each local object level and external-occlusion at the global frame level are applied using the Hidden Point Removal~(HPR) algorithm that is computationally efficient. HPR is also used for adaptively controlling the point density of each object according to the object's distance from the LiDAR. Experiment results show that the proposed \textsc{DR.CPO} algorithm is data-efficient and model-agnostic without incurring any computational overhead. Also, \textsc{DR.CPO} can improve mAP performance by 2.08\% when compared to the best 3D detection result known for KITTI dataset. The code is available at \href{https://github.com/SNU-DRL/DRCPO.git}{https://github.com/jungwook-shin/DRCPO.git}
\end{abstract}

\section{Introduction}
\label{sec:intro}
\begin{figure*}[t!]
    \centering
    \subfloat[Iterative Construction]{
    \includegraphics[width=\textwidth]{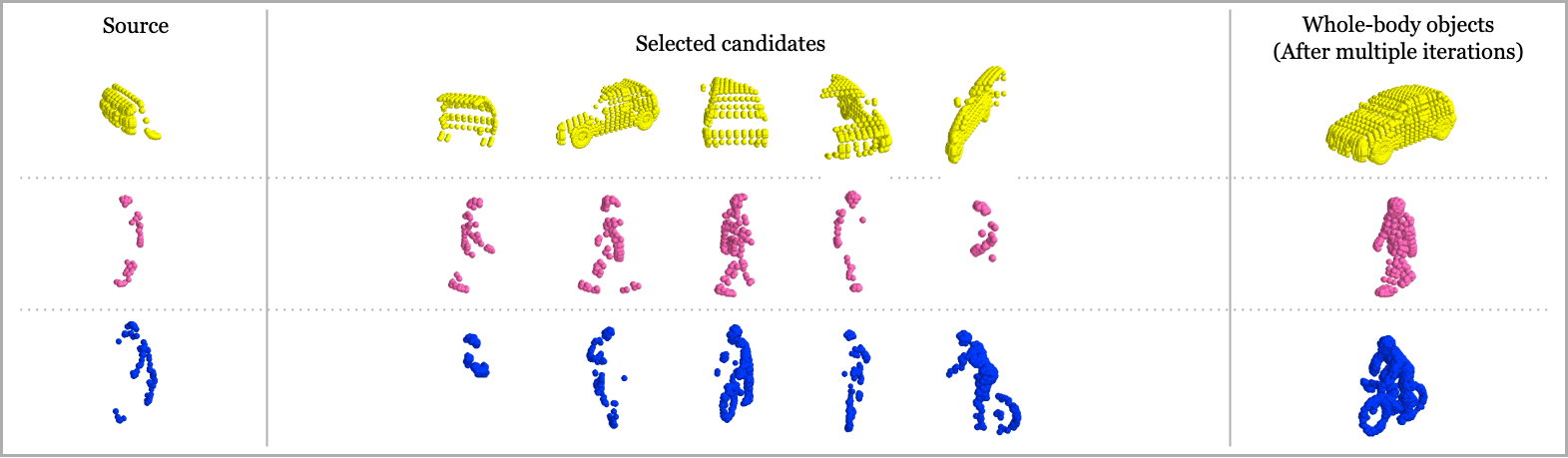}
    } \\ \vspace{0.1cm}
    \subfloat[Random Placement]{
    \includegraphics[width=0.33\textwidth]{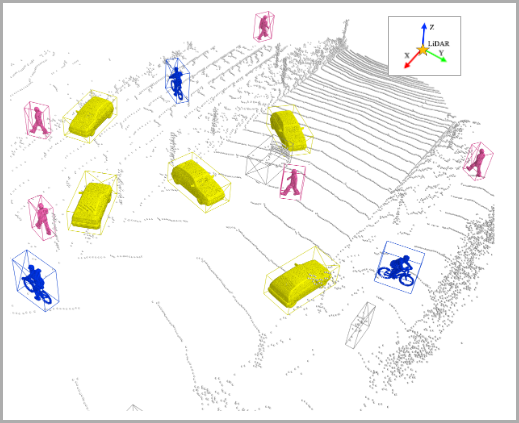}}
    \subfloat[HPR Occlusion (s-HPR)]{
    \includegraphics[width=0.33\textwidth]{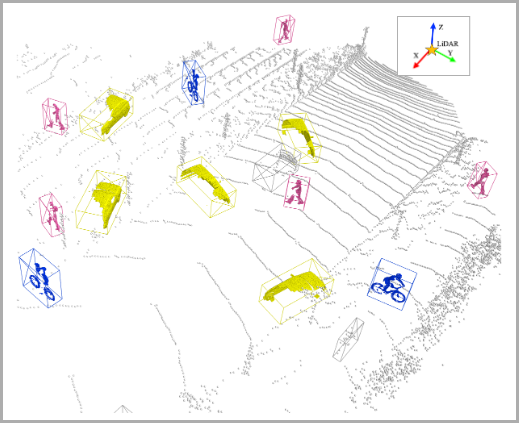}}
    \subfloat[HPR Occlusion (e-HPR)]{
    \includegraphics[width=0.33\textwidth]{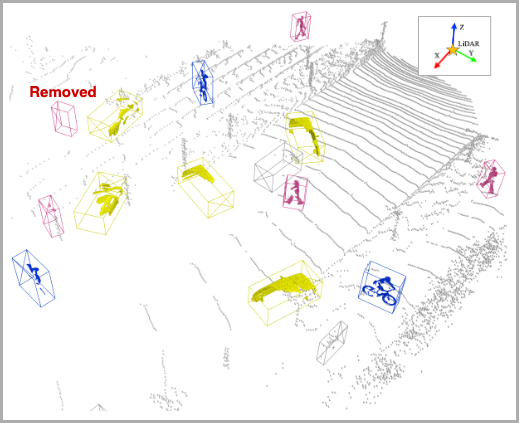}}
    \vspace{0.1cm}
    \caption{Illustration of \textsc{DR.CPO} with 3D modeling objects: 
    \label{fig:overall}
\textbf{(a) Iterative construction}: \textsc{DR.CPO} constructs a whole-body 3D object via an iterative process where each source's completion candidates are pre-selected and indexed before the model training starts.
\textbf{(b) Random placement}: The constructed whole-body object can be randomly located anywhere in the frame with a random rotation as long as its bounding box does not overlap with an existing object.
\textbf{(c, d) HPR occlusion}:
(c) s-HPR: For each individual object, a self-occlusion is applied via HPR where the distance and visibility from the fixed LiDAR viewpoint are reflected for the randomly chosen location and rotation.
(d) e-HPR: For the entire frame, an external-occlusion is applied via HPR where the inter-object spacial dependency is reflected. Some of the objects can be completely removed.
}
\vspace{0.1cm}
\end{figure*}

3D object detection using point cloud has become an important research topic~\cite{zhou2018voxelnet,he2020structure,shi2020point,yang2018pixor,pan20213d}, especially in the field of autonomous driving~\cite{geiger2013vision}. While there has been a large improvement in designing network backbone~\cite{zhou2018voxelnet,lang2019pointpillars,yan2018second,shi2019part,shi2020pv,shi2021pv} and detection head~\cite{yin2021center,zheng2021cia,hu2022afdetv2}, there has been relatively less effort on devising augmentation methods specific to the point cloud~\cite{fang2020augmented,fang2021lidar,hu2020you}.

In 3D object detection, most of the models utilize global data augmentation that is a conventional method applied to all points in a valid point cloud range.
Oversampling~\cite{yan2018second} is another conventional method that has been suggested to alleviate the extreme imbalance between the foreground and background classes. Although global augmentation and oversampling can improve the detection performance, they are straightforward methods that do not consider enhancing object-level diversity.

There have been efforts to provide object-level diversity by adopting 2D image augmentation methods such as random rotation and translation~\cite{zhou2018voxelnet,lang2019pointpillars,engelcke2017vote3deep,wang2019scnet,yang2019std}. 
The domain gap between 2D image and 3D point cloud, however, is significant and the 2D augmentation methods do not fully utilize the properties specific to 3D~\cite{choi2021part}.
Therefore, it is crucial to focus on the distinct properties of point-cloud and develop specialized augmentation methods.

Point clouds have special properties that are distinct from those of images. Most importantly, point clouds exhibit complex \textit{occlusion} because the viewpoint from LiDAR is fixed. For images, occluded pixels can be easily found by correspondence. For point clouds, there are two obstacles for manipulating occlusions~\cite{xu2022behind}. At a local object level, one part of the object is occluded by another part of the same object~(\textit{self-occlusion}). At the global level, the limitation can be extended to the entire frame where a given object can be occluded by another object~(\textit{external-occlusion}). Other properties of point cloud include \textit{point density} and \textit{intensity}. In point cloud, the number of points per a unit volume is dependent on the distance to LiDAR and each point has its own intensity which is not easy to simulate~\cite{yue2018lidar}.

To deal with the properties of 3D point cloud, we propose \textsc{DR.CPO} (\textbf{D}iversified and \textbf{R}ealistic 3D augmentation via iterative \textbf{C}onstruction, random \textbf{P}lacement, and HPR \textbf{O}cclusion). It can generate infinitely diverse frames with diversified objects, placements, and occlusions without requiring the expense of additional data collection or computational overhead. \textsc{DR.CPO} consists of three steps: (1) iterative construction, (2) random placement, and (3) HPR~(Hidden Point Removal) occlusion.
First, we generate a whole-body object in a stochastic manner using the objects from the real point cloud's ground-truth database. 
Second, we randomly rotate and place objects to anywhere inside the LiDAR detection range. Thirdly, we apply self-occlusion and external-occlusion to generate realistic views based on HPR algorithm~\cite{katz2007direct,mehra2010visibility}. Applying self-occlusion not only removes the invisible parts of an object from the LiDAR, but also controls the point density depending on the distance from the LiDAR.
Then, applying external-occlusion affects the entire frame considering the inter-object spatial dependency.
The process of \textsc{DR.CPO} is illustrated in Figure~\ref{fig:overall}, and our contributions can be summarized as below.

\begin{itemize}
    \item We introduce \textsc{DR.CPO}, a data augmentation algorithm specifically developed for 3D point clouds, to provide diversified and realistic views at the level of individual objects and at the level of entire frame.
    \item We develop an iterative method for constructing realistic whole-body objects where only real observations are used.
    \item We present a computationally light occlusion method based on HPR. Our method can implement self-occlusion, external-occlusion, and density adjustment. 
    \item We show that \textsc{DR.CPO} is data efficient, model-agnostic, and computationally efficient.
    \item We achieve state-of-the-art performance on the KITTI dataset. \textsc{DR.CPO} improves mAP performance by 2.08\%.
\end{itemize}

\section{Related Works}
\label{sec:related_works}
\vspace{0.1cm}
\paragraph{3D Object Detection: }
In voxel-based methods, VoxelNet~\cite{zhou2018voxelnet} divides point clouds into voxels where linear network, such as PointNet~\cite{qi2017pointnet}, is then applied to convert the voxels into 3D tensors.
By adopting a method of sparse convolution, SECOND~\cite{yan2018second} alleviated the computational challenges in VoxelNet, and hence improved the speed.
In point-based methods, Point-RCNN~\cite{shi2019pointrcnn} directly generates 3D proposals from raw points in a bottom-up manner and, in the second stage, the proposals are refined by combining local spatial features and global semantic features learned in the first stage.
In voxel+point based methods, PV-RCNN~\cite{shi2020pv} implements a method incorporating both voxel-based operation to generate 3D proposals and point-based operation to refine the proposals.
By implementing an advanced detection head on top of PV-RCNN, PV-RCNN++~\cite{shi2021pv} has become an improved framework that is more than 2x faster and requires less memory than PV-RCNN while achieving a comparable or better performance. Considering the advantages of PV-RCNN++, we use PV-RCNN++ as the default backbone in this work.

\vspace{0.1cm}
\paragraph{Conventional 3D Augmentations: }
Among the 3D data augmentation methods that are mostly extended from 2D augmentations (e.g., translation, rotation, scaling, flipping, and removal~\cite{qi2018frustum, wang2019frustum, shi2020pv, chen2020object, zhu2019class, hahner2020quantifying, reuse2021ambiguity}),
ground-truth sampling (GTS)~\cite{yan2018second, lehner2019patch} is a simple copy-and-paste augmentation strategy and has become popular despite its lack of consideration of occlusion between objects.
Global data augmentation (GDA)~\cite{zhou2018voxelnet, hu2022afdetv2} manipulates the entire frame globally and is another effective augmentation strategy~\cite{reuse2021ambiguity}.
After the combination of GTS and GDA was proposed as a default data augmentation set in~\cite{yan2018second}, the combination was widely employed in the subsequent studies~\cite{lang2019pointpillars, shi2019pointrcnn, shi2020pv, he2020structure}.
In our work, we refer to the augmentation that consists of GTS and GDA as the Conventional Data Augmentation~(CDA).

\vspace{0.1cm}
\paragraph{Constructing Whole-body Object: }
Learning-based approaches~\cite{tchapmi2019topnet,yang2018foldingnet,xie2020grnet} used a 3D modeling dataset including a full shape, such as Shapenet~\cite{chang2015shapenet}. However, all objects in autonomous driving dataset have incomplete shapes, including occlusions and signal misses, making a learning-based approach challenging. To use learning-based approaches, a new dataset is required for training.
Rendering-based approaches~\cite{fang2020augmented} attempted to augment 3D objects using CAD models; however, the intensity attribute was ignored in their experiments. \citet{fang2021lidar} constructed shapes using raycasting and rendering based on the depth map of 3D CAD model. In these rendering-based approaches, as with data labeling, CAD models must be manually prepared.
Geometry-based approaches~\cite{xu2022behind} constructed an object using a heuristic score that was calculated by bounding box similarity, Chamfer distance, and overlapped voxel quantity. Inspired by this, we construct a whole-body using only geometric information without the use of external data.

\vspace{0.1cm}
\paragraph{Occlusion-based Augmentation:}
To consider the shape miss in LiDAR data caused by external-occlusion, signal miss, and self-occlusion~\cite{xu2022behind}, a few works have proposed occlusion-aware augmentation methods.
In \cite{hu2020you}, the destroyed information about visibility in LiDAR data was recovered by raycasting algorithm.
When inserting virtual objects for augmentation, a visibility constraint was used to ensure that the objects are placed where they should not be occluded.
The LiDAR rendering technique automatically generates virtual objects that satisfy the occlusion constraint~\cite{abu2018augmented,fang2020augmented}. Recently, \cite{fang2021lidar} rendered virtual objects with occlusion constraint and placed them in reasonable locations of real background frames using ValidMap.
In our work, because we enforce self and external occlusion via HPR after placing virtual objects, they can be located anywhere without the need of concerning visibility or ValidMap.

\section{Methods}
\label{sec:methods}

\begin{figure}[t!]
\centering
\includegraphics[width=\columnwidth]{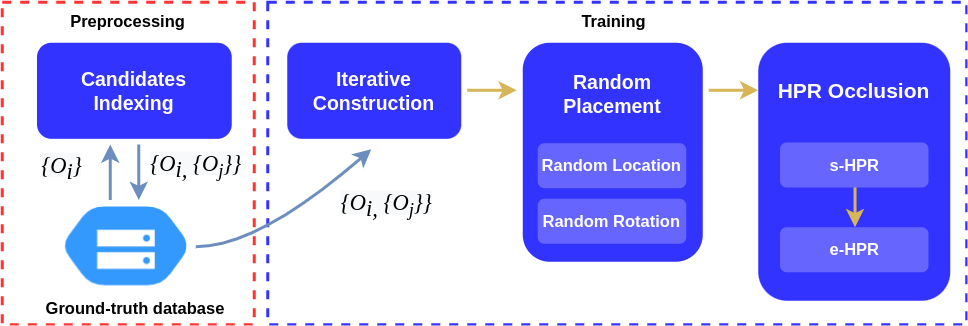}
\caption{\label{fig:augmentation_pipe}
Pipeline of \textsc{DR.CPO}: In the pre-processing phase, candidates for completion are selected for each source 3D object. The chosen candidates are indexed and additionally stored in the annotated ground-truth database such that the iterative construction step can be computed efficiently. Random placement and HPR occlusion follow the iterative construction step.}
\vspace{0.1cm}
\end{figure}

\textsc{DR.CPO} aims to generate diversified and realistic augmentations for improving 3D object detection. The pipeline of \textsc{DR.CPO} is shown in Figure~\ref{fig:augmentation_pipe} where it consists of a pre-processing step and three training-time steps.
Before presenting the details, we briefly address the notations.
A point cloud~($PC$) of a frame can be considered as the union of foreground points~($FP$) and background points~($BP$), and it can be formulated as below:
\begin{equation}
    {PC} = {FP} \cup {BP}, \ FP = \cup_{i=1}^N O_i.
\end{equation}
In the equation, $O_i$ is the $i$'th object in the frame and its information can be expressed as $(P_i, s_i, B_i)$.
\begin{itemize}
\item{$P_i$ refers to a set of points, $P_i=\left\{ p_{i,1}, \dots, p_{i,T_{i}} \right\}$, where each point $p_{i,t}$ is a 4-dimensional vector $(x, y, z, r)$ with the first three as the 3D coordinates and $r$ as the point's reflection intensity.}
\item{$s_i$ is the ground-truth class label with three possible classes: $s_i\in\left\{ \text{Car}, \text{Pedestrian}, \text{Cyclist} \right\}$.} 
\item{$B_i$ contains the annotation of $O_i$'s 3D bounding box. Specifically, $B_i=(\text{cx}_i, \text{cy}_i, \text{cz}_i, l_i, w_i, h_i, \theta_i)$ with the first three as the center's coordinates, the following three as the length, width, and height of the bounding box, and $\theta_i$ as the heading angle of the bounding box.}
\end{itemize}

\subsection{Iterative Construction}
The goal of this step is to sample a source object $O_i$ from the ground-truth database and to construct it back to a whole-body object. To do so, we select another object $O_j$ that has the same class label ($s_i=s_j$) and can help fill up the missing points of $O_i$. Then, $P_i$ is complemented by additionally including the points $P_j$ into $P_i$. This complementing operation is iteratively repeated until $O_i$ is sufficiently complemented to have $P_i$ represent a whole-body object. An illustration of the iterative construction can be found in Figure~\ref{fig:overall}(a) and visualizations for KITTI can be found in the supplementary\footnote{Refer to our arXiv version for the Supplementary materials.}~(Figure 8, 9, and 10
). For a given source object $O_i$, we maximize the diversity of the constructed object by making the iterative construction stochastic and by performing the iterative construction on the fly during the training. The computational overhead is minimized by performing an one-time pre-processing of the ground-truth database where the set of desirable candidates for complementing, $\{O_j\}$, are indexed for each source object $O_i$.

\vspace{0.1cm}
\subsubsection{Indexing Candidate Objects~(Pre-processing):}
In this pre-processing step, we identify and index $K$ objects as the candidates for complementing each object $O_i$ in the database. An object $O_j$ is considered to be a candidate for complementing $O_i$ if it satisfies two criteria: (1) high bounding box size similarity, and (2) high partition density for the low density partitions of $O_i$.
First, we convert all the objects into a canonical pose such that they have the same heading angle (i.e., $\theta_i=0$ for all $i$).
Then, we calculate the bounding box similarity with $O_i$ for all the objects $O_j$~($i\neq j$) in the ground-truth database to select the best $2K$ candidates. Among the $2K$ candidates, the best $K$ candidates that have high density partitions for $O_i$'s low density partitions (due to occlusion, signal miss, sparsity, etc.) are selected as the final candidates. Note that we define the partitions as the non-overlapping subdivided regions of the object $O_i$ with a fixed number of partitions for each class.
Also, we evaluate the density as the number of points per partition.
Further details can be found in Supplementary A.1.
$K=400$ was used for our experiments.
Partition can be considered as an extension of voxelization. In this study, we adopt the idea of partition to make it easy to analyze the density of each partition. This allows an effective method for combining multiple objects. Previously, partition was used in~\cite{choi2021part} for the purpose of applying a different augmentation technique to each partition.

\vspace{0.1cm}
\subsubsection{Constructing a Whole-body Object (Training Time):}
For a randomly selected source object $O_i$, we iteratively and stochastically construct its whole-body using the indexed $K$ objects. The details can be summarized as the following where all steps are performed in the canonical pose.
Since the car and cyclist objects are (roughly) symmetric on the x-axis in the canonical pose, (1) we first mirror their points as an additional operation before the iterative complementing. This step is skipped for the pedestrian objects.
Then, (2) for each partition, we uniform randomly sample a candidate from the $K$ candidates and add its points to $P_i$. 
(3) We iteratively repeat (2) until the object $O_i$ can be regarded as a whole-body object. We determine the object to be a whole-body when the proportion of partitions whose density is higher than the mean is at least 85\%. For a formal description and examples, see Supplementary A.1.
Due to the stochastic nature of candidate sampling in (2), our method can generate \textit{combinatorially diverse} whole-body objects for a given source object $O_i$.

\begin{figure}[t!]
    \centering
    \includegraphics[width=0.49\textwidth]{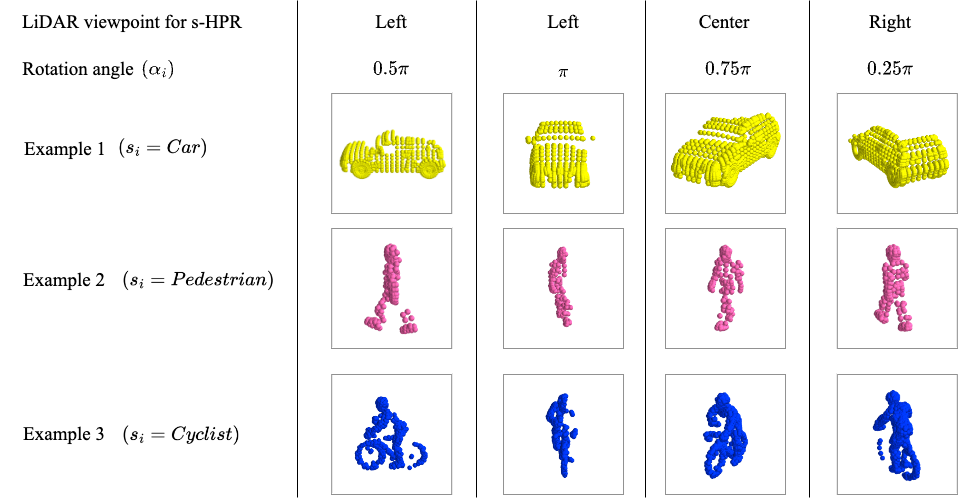}
    \caption{\label{fig:diversity}Diversification via placement: An iteratively constructed whole-body object can be placed at a random location, resulting in a random distance and a random viewpoint from the LiDAR. Therefore, the randomness in location provides a dimension of diversification. Also, the object can be randomly rotated and provide another dimension of diversification. Subsequently, s-HPR is applied according to the location and rotation to make the augmented point-cloud objects look realistic. The three objects constructed in Figure~\ref{fig:overall}(b) were used for this illustration.} 
\end{figure}

\subsection{Random Placement}
In the conventional data augmentation, an object from the ground-truth database must be placed at the original location with the original angle. If not, the occlusion and density of the object does not match the placement and thus causing a negative effect for the learning. \textsc{DR.CPO} does not suffer from this limitation because an object from ground-truth database is first transformed into a whole-body object and because a proper occlusion and density adjustment for the chosen location and angle are reflected in the HPR occlusion step. Taking full advantage of the flexibility, we place each whole-body object randomly and maximize the diversity of the augmentation as in the example of Figure~\ref{fig:overall}(b). To be specific, we first randomly rotate the object along the z-axis:  $\theta_i=\alpha_i$, where $\alpha_i$ is uniformly sampled between $-1.00\pi$ and $+1.00\pi$. Then, we select the location following a uniform distribution over the $(x,y)$ coordinates. 
As we will show later, the diversity shown in Figure~\ref{fig:diversity} through the randomness is essential for improving the performance and enhancing the data efficiency.

\vspace{0.2cm}
\subsection{HPR Occlusion}
To enforce self-occlusion and external-occlusion, we utilize Hidden Point Removal (HPR)~\cite{katz2007direct}.
We have chosen HPR mainly because of its computational efficiency. To the best of our knowledge, we are the first to adopt HPR for enforcing occlusion in autonomous driving.

\vspace{0.1cm}
\subsubsection{Hidden Point Removal (HPR):}
Given a set of points, $P_i$, our goal is to determine if each point $p_{i,t}$ is visible from the LiDAR viewpoint $C$. For this purpose, we use the HPR method illustrated in Figure 11
in the supplementary.
To determine whether the point $p_{i,t}$ is visible from $C$, HPR defines a sphere with a sufficiently large radius $R$ centered at origin $C$.
Spherical flipping is used to reflect every point $p_{i,t}$ internal to the sphere along the ray from $C$ to $p_{i,t}$ to its image outside the sphere, following the equation:
\begin{equation}
    \hat{p}_{i,t} = p_{i,t} + 2(R-||p_{i,t}||) \frac{p_{i,t}}{||p_{i,t}||}
\end{equation}
where $||p_{i,t}||$ is the length of $p_{i,t}$ from $C$. If we define $\hat{P}_i=\left\{ \hat{p}_{i,t} \right\}_{t=1}^{T}$, the visibility of $p_{i,t}$ is determined by whether it lies on the convex hull of $\hat{P}_{i}\cup \left\{ C \right\}$. Rigorously speaking, we choose $\epsilon$-visible points by adjusting $R$ for each class. Thanks to the efficient convex hull operation, HPR is computationally outstanding while meeting the precision required for the LiDAR data augmentation. For a full description of HPR, see \cite{katz2007direct}.

\vspace{0.1cm}
\subsubsection{s-HPR and e-HPR for Occlusion:}
\begin{figure}[t!]
\centering
\includegraphics[width=\columnwidth]{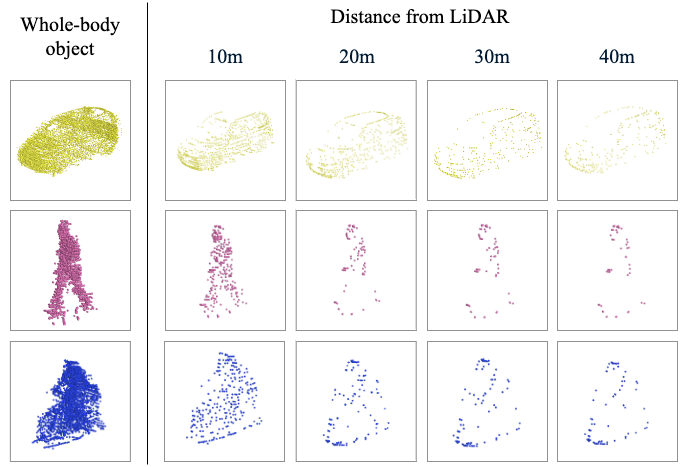}
\caption{\label{fig:sparsity_with_distance} 
Density adjustment with s-HPR: Depending on the object's distance from the LiDAR, point density is adjusted by s-HPR. Three examples are shown for the whole-body objects constructed from KITTI dataset.
}
\vspace{-0.1cm}
\end{figure}

We first apply HPR to perform self-occlusion~(s-HPR) and subsequently to perform external-occlusion~(e-HPR) as shown in Figure~\ref{fig:overall}(c) and \ref{fig:overall}(d). For s-HPR, HPR is applied to each individual object where the object's location and angle are accounted for. A crucial advantage of using s-HPR is that it automatically adjusts the point density of the object depending on the distance from the LiDAR. Examples are shown in Figure~\ref{fig:sparsity_with_distance}. The density adjustment is an important feature to make the augmented object's occlusion look realistic, and its effect on performance will be addressed in Section~\ref{subsec:hpr_occlusion}. Another important role of s-HPR is to reduce the number of points such that the computational burden of e-HPR can be lessened. For e-HPR, HPR is applied to the entire frame such that external-occlusion related to the inter-object dependency can be calculated. Compared to the s-HPR, this is a computationally heavy operation and the reduction in the number of points by s-HPR is beneficial for keeping e-HPR's computation burden low. Computational efficiency of \textsc{DR.CPO} will be addressed in Section~\ref{subsec:comp_efficiency}.

\section{Experiments}
\label{sec:experiments}

In this section, we show that \textsc{DR.CPO} is a data-efficient and model-agnostic augmentation method that can improve the performance of voxel, point, and voxel+point models, including the state-of-the-art voxel+point model, without incurring an extra computational cost.
For empirical evaluations, we use KITTI~\cite{geiger2013vision} dataset that is a standard 3D LiDAR point cloud benchmark and evaluate the standard metric $AP_{R40}$.
For a fair comparison, \textsc{DR.CPO} is set to follow the same class distribution as in CDA (details can be found in Table~\ref{tab:speed_comparisons}).
Unless stated otherwise, we use PVRCNN++~\cite{shi2021pv}, an open-source and state-of-the-art voxel+point method, as our default backbone model.
For training any model with \textsc{DR.CPO}, we trained the model by following the original protocol and code.
Additional details of the implementations and experimental settings are provided in Supplementary B.

\subsection{Data Efficiency}
\label{subsec:data-efficiency}
\begin{figure}[t!]
\centering
  \centering
  \subfloat{
    \includegraphics[width=0.5\columnwidth]{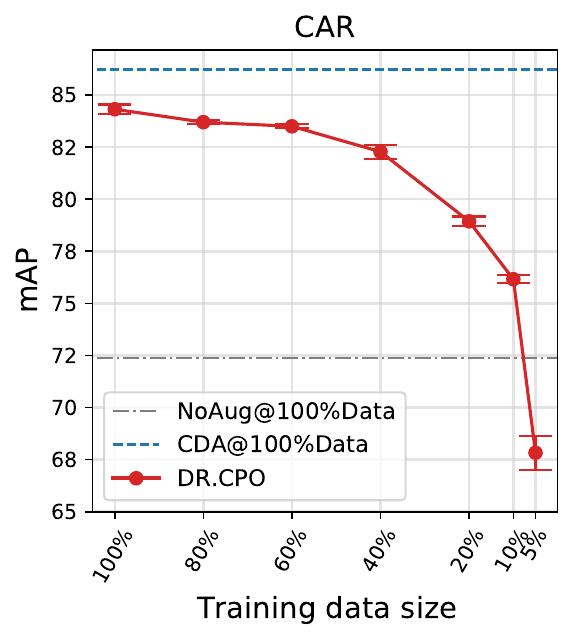}
  }
  \subfloat{
    \includegraphics[width=0.5\columnwidth]{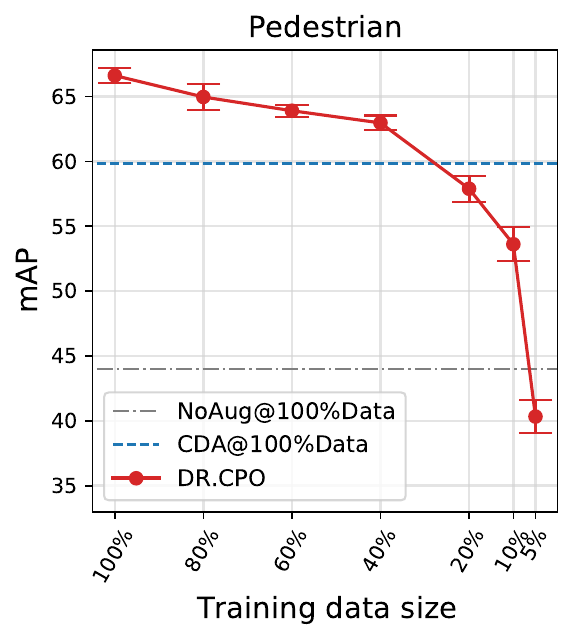}
  }
  \textbf{}
  \subfloat{
    \includegraphics[width=0.5\columnwidth]{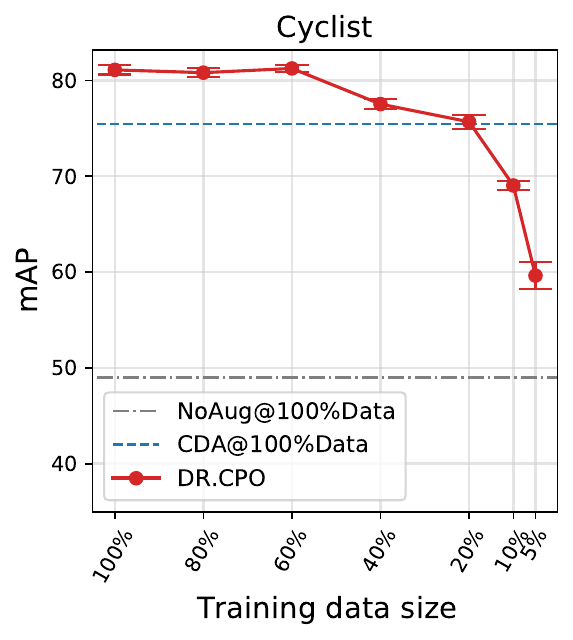}
  }
  \subfloat{
    \includegraphics[width=0.5\columnwidth]{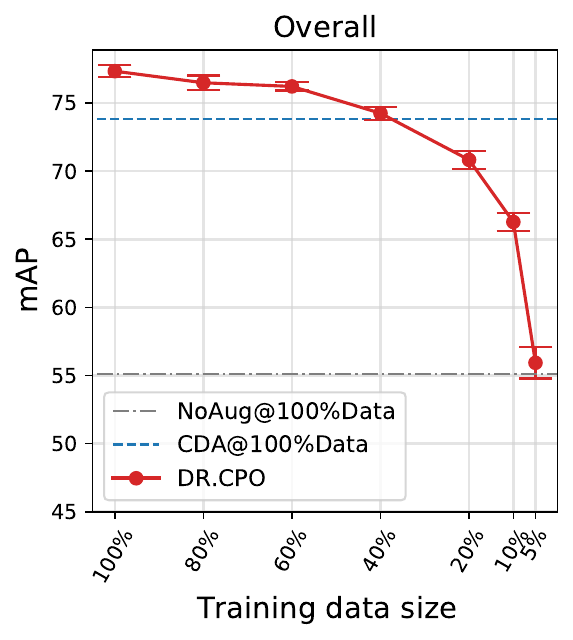}
  }
\caption{\label{fig:data_efficiency}
Data efficiency of \textsc{DR.CPO}: The training data size versus mAP score is plotted for KITTI. Except for car, \textsc{DR.CPO} achieves significant improvements over no-augmentation and CDA.}
\end{figure}

To examine the data efficiency of \textsc{DR.CPO}, we trained our model using subsets of KITTI's training examples. The results are shown in Figure~\ref{fig:data_efficiency}. Compared to the baseline (no augmentation) and CDA shown in gray and blue dotted lines, respectively, \textsc{DR.CPO} can match their performance with a much smaller data size. An exception is the class of car for CDA, and this is discussed in Section~\ref{sec:discussion}. For the overall performance, \textsc{DR.CPO} can perform as well as the baseline and CDA using only 4.60\% and 37.60\% of the training dataset, respectively.

\subsection{Model-agnostic Effectiveness}
\begin{table}[t!]
\centering
\caption{\label{tab:model_agnostic} \textsc{DR.CPO} is a model-agnostic augmentation method. \textsc{DR.CPO} can effectively enhance the performance of existing models that are based on voxel, point, and voxel+point representations. $AP_{R40}$ performance is shown.}
\resizebox{\columnwidth}{!}{
\begin{threeparttable}[b]
\begin{tabular}{@{}lccccc@{}}
\toprule
Method & Car & Pedestrian & Cyclist & Mean & Diff. \\\midrule
SECOND\tnote{$\star$} ~\cite{yan2018second}      & 82.67   & 52.31   & 62.69   & 65.89 & $-$3.64 \\
SECOND with \textsc{DR.CPO} & \textbf{83.15}   & 51.67   & \textbf{73.75}   & \textbf{69.53} &  \\ \midrule
Point-RCNN~\cite{hu2021pattern} & 82.84   & 56.92   & 77.46   & 72.40 & $-$1.08          \\
Point-RCNN with \textsc{DR.CPO}   & 82.36   & \textbf{59.86}   & \textbf{78.22}   & \textbf{73.48} &  \\ \midrule
PV-RCNN\tnote{$\star$} ~\cite{shi2020pv}& \textbf{86.70} & 57.61 & 75.87 & 73.39 & $-$2.79 \\
PV-RCNN with \textsc{DR.CPO}& 84.75 & \textbf{62.33} & \textbf{81.47} & \textbf{76.18} &  \\  \bottomrule
\end{tabular}
\begin{tablenotes}
\small
\item[] The results marked with $\star$ are from BtcDet~\cite{xu2022behind}.
\end{tablenotes}
\end{threeparttable}
}
\vspace{-0.8cm}
\end{table}

\begin{table*}[thb!]
\centering
\caption{\label{tab:big_comparisons} 
Compilation of KITTI 3D LiDAR detection performance. State-of-the-art models are also included. For each column, any model performance that is superior to ours is indicated in bold.}
\resizebox{\textwidth}{!}{
\begin{threeparttable}
\begin{tabular}{@{}lcccccccccccccc@{}}
\toprule
\multirow{2}{*}{Method} & \multicolumn{3}{c}{Car} & \multicolumn{3}{c}{Pedestrian} & \multicolumn{3}{c}{Cyclist} & \multicolumn{3}{c}{Overall} & \multirow{2}{*}{Mean} & \multirow{2}{*}{Diff.} \\ \cmidrule(lr){2-4}\cmidrule(lr){5-7}\cmidrule(lr){8-10}\cmidrule(lr){11-13}
 & E & M & D & E & M & D & E & M & D & Car & Ped & Cyc & & \\ \midrule
ContFuse~\cite{liang2018deep}                      & 83.68          & 68.78          & 61.67          & -              & -              & -              & -              & -              & -              & 71.38          & -              & -              & -              &        \\
3D IoU Loss~\cite{zhou2019iou}                     & 84.43          & 76.28          & 68.22          & -              & -              & -              & -              & -              & -              & 76.31          & -              & -              & -              &        \\
SECOND with LiDAR-Aug~\cite{fang2021lidar}         & 88.65          & 76.97          & 70.44          & 58.53          & 54.56          & 51.78          & -              & -              & -              & 78.69          & 54.96          & -              & -              &        \\
PointPillars with LiDAR-Aug~\cite{fang2021lidar}   & 87.75          & 77.83          & 74.90          & 59.99          & 55.15          & 52.66          & -              & -              & -              & 80.16          & 55.93          & -              & -              &        \\
PointRCNN with LiDAR-Aug~\cite{fang2021lidar}      & 89.56          & 79.51          & 77.89          & 67.46          & 59.06          & 56.23          & -              & -              & -              & 82.32          & 60.92          & -              & -              &        \\
PV-RCNN with LiDAR-Aug~\cite{fang2021lidar}        & 90.18          & \textbf{84.23} & 78.95          & 65.05          & 59.90          & 55.52          & -              & -              & -              & 84.45          & 60.16          & -              & -              &        \\
Part-A2-Net~\cite{guo2020deep}                     & \textbf{91.70} & \textbf{87.79} & \textbf{84.61} & -              & -              & -              & 81.91          & 68.12          & 61.92          & \textbf{88.03} & -              & 70.65          & -              &        \\
PV-RCNN with STRL~\cite{huang2021spatio}           & -              & -              & -              & -              & -              & -              & -              & -              & -              & 84.70          & 57.80          & 71.88          & 71.46          & \phantom{0}$-$6.84  \\
VoxelNet~\cite{zhou2018voxelnet}                   & 81.97          & 65.46          & 62.85          & 57.86          & 53.42          & 48.87          & 67.17          & 47.65          & 45.11          & 70.09          & 53.38          & 53.31          & 58.93          & $-$19.37 \\
StarNet~\cite{ngiam2019starnet}                    & 81.63          & 73.99          & 67.07          & 48.58          & 41.25          & 39.66          & 73.14          & 58.29          & 52.58          & 74.23          & 43.16          & 61.34          & 59.58          & $-$18.72 \\
PointPillars with PA-AUG~\cite{choi2021part}          & 83.70          & 72.48          & 68.23          & 57.38          & 51.85          & 46.91          & 70.88          & 47.58          & 44.80          & 74.80          & 52.05          & 54.42          & 60.42          & $-$17.87 \\
PV-RCNN with PA-AUG~\cite{choi2021part}            & 89.38          & 80.90          & 78.95          & 67.57          & 60.61          & 56.58          & 86.56          & 72.21          & 68.01          & 83.08          & 61.59          & 75.59          & 73.42          & \phantom{0}$-$4.88  \\
StarNet with PPBA~\cite{cheng2020improving}        & 84.16          & 77.65          & 71.21          & 52.65          & 44.08          & 41.54          & 79.42          & 61.99          & 55.34          & 77.67          & 46.09          & 65.58          & 63.12          & $-$15.18 \\
Point-GNN~\cite{guo2020deep}                       & \textbf{93.11} & \textbf{89.17} & \textbf{83.90} & 55.36          & 47.07          & 44.61          & 81.17          & 67.28          & 59.67          & \textbf{88.73} & 49.01          & 69.37          & 69.04          & \phantom{0}$-$9.26  \\
HotSpotNet~\cite{chen2020object}                   & 87.60          & 78.31          & 73.34          & \textbf{82.59} & 65.95          & 59.00          & 53.10          & 45.37          & 41.47          & 79.75          & \textbf{69.18} & 46.65          & 65.19          & $-$13.11 \\
Object as Hotspots (Dense)~\cite{chen2020object}   & 91.09          & \textbf{82.20} & 79.69          & \textbf{85.85} & 66.45          & 62.16          & 68.88          & 62.82          & 55.78          & 84.33          & \textbf{71.49} & 62.49          & 72.77          & \phantom{0}$-$5.53  \\
STD~\cite{guo2020deep}                             & \textbf{94.74} & \textbf{89.19} & \textbf{86.42} & 60.02          & 48.72          & 44.55          & 81.36          & 67.23          & 59.35          & \textbf{90.12} & 51.10          & 69.31          & 70.18          & \phantom{0}$-$8.12  \\
3DSSD~\cite{guo2020deep}                           & \textbf{92.66} & \textbf{89.02} & \textbf{85.86} & 60.54          & 49.94          & 45.73          & 85.04          & 67.62          & 61.14          & \textbf{89.18} & 52.07          & 71.27          & 70.84          & \phantom{0}$-$7.46  \\
PointRCNN~\cite{hu2021pattern}                     & 90.10          & 80.41          & 78.00          & 64.18          & 56.71          & 49.86          & 91.72          & 72.47          & 68.18          & 82.84          & 56.92          & 77.46          & 72.40          & \phantom{0}$-$5.89  \\
PV-RCNN with Pattern-Aware GT~\cite{hu2021pattern} & \textbf{92.13} & \textbf{84.79} & \textbf{82.56} & 65.99          & 58.57          & 53.66          & 90.38          & 72.03          & 67.96          & \textbf{86.49} & 59.41          & 76.79          & 74.23          & \phantom{0}$-$4.07  \\ \midrule
SECOND\tnote{$\star$} ~\cite{yan2018second}                         & 90.97          & 79.94          & 77.09          & 58.01          & 51.88          & 47.05          & 78.50          & 56.74          & 52.83          & 82.67          & 52.31          & 62.69          & 65.89          & $-$12.41 \\
PointPillars\tnote{$\star$} ~\cite{lang2019pointpillars}                   & 87.75          & 78.39          & 75.18          & 57.30          & 51.41          & 46.87          & 81.57          & 62.94          & 58.98          & 80.44          & 51.86          & 67.83          & 66.71          & $-$11.59 \\
BtcDet~\cite{xu2022behind}                         & \textbf{93.15} & \textbf{86.28} & \textbf{83.86} & 69.39          & 61.19          & 55.86          & 91.45          & 74.70          & 70.08          & \textbf{87.76} & 62.15          & 78.74          & 76.22          & \phantom{0}$-$2.08  \\
PV-RCNN\tnote{$\star$} ~\cite{shi2020pv}                        & \textbf{92.57} & \textbf{84.83} & \textbf{82.69} & 64.26          & 56.67          & 51.91          & 88.88          & 71.95          & 66.78          & \textbf{86.70} & 57.61          & 75.87          & 73.39          & \phantom{0}$-$4.90  \\
PV-RCNN++~\cite{shi2021pv}(Reproduced)              & \textbf{91.72} & \textbf{84.82} & \textbf{82.03} & 66.00          & 60.05          & 54.70          & 91.39          & 70.14          & 66.92          & \textbf{86.19} & 60.25          & 76.15          & 74.20          & \phantom{0}$-$4.10  \\ \midrule
PV-RCNN++ with \textbf{\textsc{DR.CPO} (ours)}                                                   & 91.27          & 81.67          & 81.28          & 73.23          & \textbf{67.66} & \textbf{62.96} & \textbf{93.08} & \textbf{78.21} & \textbf{75.32} & 84.74          & 67.95          & \textbf{82.20} & \textbf{78.30} &        \\ \bottomrule
\end{tabular}
\begin{tablenotes}
  \small
  \item[] The results marked with $\star$ are from BtcDet~\cite{xu2022behind}.
\end{tablenotes}
\vspace{-0.3cm}
\end{threeparttable}
}
\end{table*}

Because \textsc{DR.CPO} is an augmentation method, we have chosen three latest and well-known models and examined if their performance can be enhanced with \textsc{DR.CPO}. The results are shown in Table~\ref{tab:model_agnostic}. In the table, SECOND is a voxel-based method, Point-RCNN is a point-based method, and PV-RCNN is a voxel+point-based method. The improvements from \textsc{DR.CPO} are 3.64\%, 1.08\%, and 2.79\%, respectively.

\subsection{Comparison with State-of-the-art Models}
Because of the improved detection head and the computationally efficient representative keypoints, we have used PV-RCNN++ as the backbone of \textsc{DR.CPO} and achieved mAP score of 78.30\%. We have compiled the previously reported records as much as we can into Table~\ref{tab:big_comparisons}, where the latest models such as SECOND, PointPillars, BtcDet, PV-RCNN, PV-RCNN++ are also included. In particular, BtcDet~\cite{xu2022behind}, which integrates multiple techniques including an extra shape occupancy network, is known to be the current state-of-the-art model. Simply by incorporating \textsc{DR.CPO} into the augmentation pipeline, the computationally efficint PV-RCNN++ can perform better than all the other models shown in Table~\ref{tab:big_comparisons} including BtcDet.
As in the data efficiency and model-agnostic effectiveness studies, the gain comes from the extraordinary improvements in pedestrian and cyclist categories. The improvements are large enough to overcome the loss in car category. We discuss the loss of car category in Section~\ref{sec:discussion}.

\subsection{Computational Efficiency}
\label{subsec:comp_efficiency}
For all of our experiments, \textsc{DR.CPO} was set to follow the same class distribution as in CDA for a fair comparison. This can be confirmed from the average object counts per frame in Table~\ref{tab:speed_comparisons}. Even with similar object counts, \textsc{DR.CPO} needs to perform additional computations including iterative construction and HPR occlusion. The computational efficiency in terms of training time per epoch, however, is not compromised as can be seen in the same table. In fact, \textsc{DR.CPO} is slightly faster than CDA. Despite the overhead of our augmentation method, the training time is more important than the data preparation time and \textsc{DR.CPO} requires less training time thanks to the reduced number of points after s-HPR and e-HPR. Overall, \textsc{DR.CPO} does not incur any additional burden when compared to the ubiquitously used CDA.

\begin{table}[t!]
\caption{\label{tab:speed_comparisons}
Object counts, the number of points, and training time per epoch.
Our method increases objects per frame without increasing the total points due to s-HPR and e-HPR operations. Overall training time does not increase because the delay in time on data preparation is offset by reduced time on processing model.
}
\centering
\resizebox{\columnwidth}{!}{
\begin{tabular}{@{}llccc@{}}
\toprule
                                &                       & No Aug. & CDA   & Ours  \\ \midrule
Average counts per frame        & Car objects           & \phantom{0.}4.00    & \phantom{.}13.70  & \phantom{.}13.40  \\
                                & Pedestrian objects    & \phantom{0.}0.60    & \phantom{.}11.80  & \phantom{.}12.90  \\
                                & Cyclist objects       & \phantom{0.}0.20    & \phantom{.}10.30  & \phantom{.}11.00  \\
Total points per frame          &                       & 18912  & 23170 & 17203 \\ \midrule
Training time per epoch (sec)   & On data preparation   & \phantom{00}2.10    & \phantom{00}4.11   & \phantom{00}9.55  \\
                                & On processing model   & 124.10  & 134.26 & 117.21 \\
                                & Total                 & 126.10  & 138.37 & 129.40 \\ \midrule
Performance (mAP)               &                       & \phantom{0}55.06  & \phantom{0}73.48 & \phantom{0}78.30 \\ \bottomrule
\end{tabular}
}
\vspace{0.1cm}
\end{table}

\section{Analysis of \textsc{DR.CPO}}
\label{sec:analysis}
In this section, we analyze \textsc{DR.CPO}'s three elements. When analyzing 
an element, the other two are kept the same as in the default setup of \textsc{DR.CPO}. 

\subsection{Iterative Construction}
\label{subsec:iter_const}
We have examined if constructing a whole-body object can indeed positively influence the performance, and the results are shown in Table~\ref{tab:construction}. By looking at the mean performance column, it can be confirmed that self-mirroring and iteration for complementing are both helpful. The gain from mirroring is only 0.11\% (the difference between the first and second rows). The performance gain is mostly due to the iteration process, and it is 95\% of the total gain.
From the car column, it can be seen that multiple iterations can actually hurt the performance of the car detection. 
For the cyclist column, however, the gain from single iteration to multiple iteration is large (1.63\%).

\subsection{Random Placement}
To analyze the effects of random location and random rotation, we have performed an ablation study and the results can be found in Table~\ref{tab:analysis_placement1}. While both are helpful as expected, the random location has improved the mean performance by 0.22\% while random rotation has improved it by 1.51\%. Apparently, the additional diversity from rotation is more effective than the additional diversity from random location. To further investigate random rotation, we have analyzed the effect of rotation range and the results are shown in 
Table~\ref{tab:analysis_placement2}. It can be observed that generally a larger rotation range is better. It is interesting to note that the largest gain is obtained when the rotation range is increased from $(-0.75\pi, +0.75\pi)$ to $(-1.00\pi, +1.00\pi)$, where the gain is 1.02\%. Rotation of $1.00\pi$ corresponds to switching the front and the back sides of an object, and thus inverting the direction of movement. 

\begin{table}[t!]
\caption{\label{tab:construction}Iterative construction: the effect of mirroring and iteration.}
\centering
\resizebox{\columnwidth}{!}{
\begin{tabular}{@{}llccccc@{}}
\toprule
Mirroring & \begin{tabular}[c]{@{}l@{}}Iteration\end{tabular} & Car & Pedestrian & Cyclist & Mean & Diff.  \\ \midrule
No & None & 85.08   & 63.01   & 79.75   & 75.95 & $-$2.35 \\
Yes & None & 85.18 & 63.31 & 79.65 & 76.06 & $-$2.24 \\
Yes & Single & 85.25 & 67.85 & 80.57 & 77.89 & $-$0.41        \\
Yes & Multiple & 84.74 & 67.95 & 82.20 & 78.30 &         \\ \bottomrule
\end{tabular}
}
\vspace{0.1cm}
\end{table}

\begin{table}[t!]
\caption{\label{tab:analysis_placement1}Random placement: ablation study of location and rotation.}
\centering
\resizebox{\columnwidth}{!}{
\begin{tabular}{@{}cccccccc@{}}
\toprule
    \begin{tabular}[c]{@{}l@{}}Random \\ Location\end{tabular}  & \begin{tabular}[c]{@{}l@{}}Random \\ Rotation\end{tabular}   & Car & Pedestrian      & Cyclist   & Mean & Diff. \\ \midrule
                    &                   & 84.31 &  65.58        & 79.85     & 76.58 & $-$1.72 \\
     \checkmark     &                   & 84.28 &  67.01        & 79.32     & 76.87 & $-$1.43 \\ 
                    & \checkmark        & 85.33 &  66.00        & 83.16     & 78.16 & $-$0.14 \\
     \checkmark     & \checkmark        & 84.74 &  67.95        & 82.20     & 78.30 &  \\
    \bottomrule
\end{tabular}
}
\end{table}

\begin{table}[t!]
\caption{\label{tab:analysis_placement2}Random placement: the effect of rotation range.}
\centering
\resizebox{\columnwidth}{!}{
\begin{tabular}{@{}ccccccc@{}}
\toprule
    Rotation range  & Car & Pedestrian & Cyclist   & Mean & Diff.\\ \midrule
    $(-0.25\pi, +0.25\pi)$ & 85.13 & 64.41 & 80.87 & 76.81 & $-$1.49 \\
    $(-0.50\pi, +0.50\pi)$ & 85.36 & 64.99 & 80.46 & 76.94 & $-$1.36 \\ 
    $(-0.75\pi, +0.75\pi)$ & 84.56 & 65.67 & 81.61 & 77.28 & $-$1.02 \\
    $(-1.00\pi, +1.00\pi)$ & 84.74 & 67.95 & 82.20 & 78.30 &  \\
    \bottomrule
\end{tabular}
}
\end{table}

\begin{table}[t!]
\caption{\label{tab:hpr_ablation}HPR occlusion: ablation study of s-HPR and e-HPR. s-HPR is responsible for density adjustment in addition to self-occlusion.}
\centering
\resizebox{\columnwidth}{!}{
\begin{tabular}{@{}cccccccc@{}}
\toprule
    s-HPR       & e-HPR          & Car   & Pedestrian  & Cyclist  & Mean & Diff. \\ \midrule
                &               & 84.64 & 52.33 & 78.94 & 71.97 & $-$6.33 \\
                & \checkmark    & 84.98 & 64.83 & 80.22 & 76.68 & $-$1.62 \\
     \checkmark &               & 85.47 & 63.58 & 83.05 & 77.37 & $-$0.93 \\
     \checkmark & \checkmark    & 84.74 & 67.95 & 82.20 & 78.30 & \\ 
    \bottomrule
\end{tabular}
}
\end{table}

\subsection{HPR Occlusion}
\label{subsec:hpr_occlusion}

To understand the effects of s-HPR and e-HPR, an ablation study was performed and the results are shown in Table~\ref{tab:hpr_ablation}. While the effectiveness of both can be confirmed, the impact is larger when s-HPR is removed (-1.62\%) than when e-HPR is removed (-0.93\%). This can be explained by s-HPR's additional functionality of adjusting density as explained in Figure~\ref{fig:sparsity_with_distance}.
When both s-HPR and e-HPR are turned off, the mean performance becomes 71.97\% and it is even worse than the CDA baseline of 74.20\% that can be found in Table~\ref{tab:big_comparisons} (mean performance of PV-RCNN++). This indicates that it is indeed undesirable to augment with whole-body objects and do not reflect the effects of occlusion. An augmented object becomes unrealistic without a proper occlusion, and it is better to avoid iterative construction and random placement when the occlusion step does not exist.

\begin{figure}[t!]
\centering
\includegraphics[width=\columnwidth]{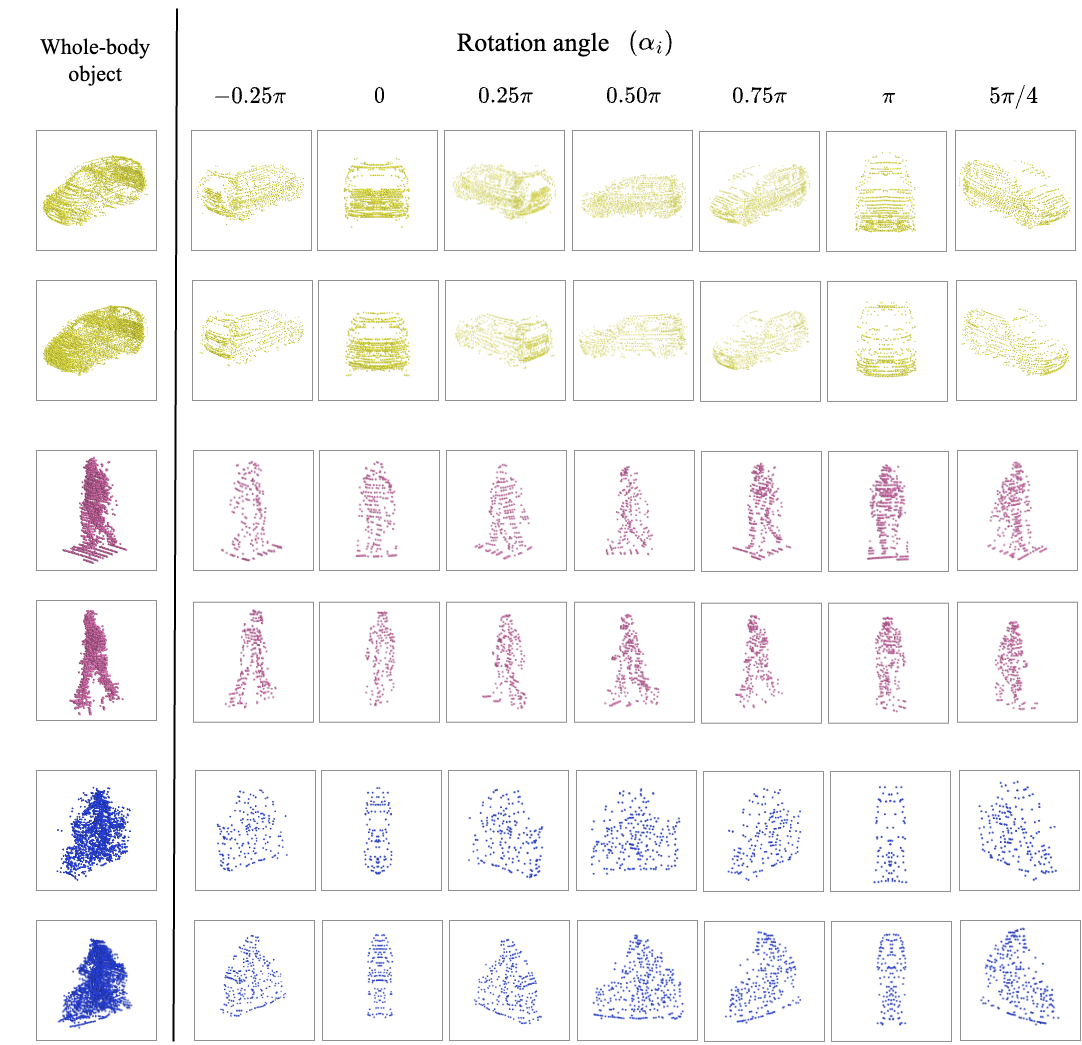}
\caption{\label{fig:diversity_rotation}
Diversity of the occluded shapes: Applying s-HPR to the randomly rotated objects results in a wide range of shapes depending on the object's rotation angle. For the above examples, distance is fixed at 10m and LiDAR viewpoint is always at the center.}
\end{figure}

\section{Discussion}
\label{sec:discussion}

\paragraph{Diversified and Realistic Augmentation by \textsc{DR.CPO}:}
The conventional data augmentation method is constrained in terms of the augmentation diversity it can provide - it relies on insertion of copied objects and rotation and scaling of the entire training frame. Compared to CDA, our method can provide diversified augmentations mainly via combinatorially stochastic nature of whole-body construction, random location, and random rotation. Their effectiveness for enhancing performance has been shown in Table~\ref{tab:construction}, \ref{tab:analysis_placement1}, and  \ref{tab:analysis_placement2}. 
Illustrations of diversification via placement can be found in  Figure~\ref{fig:diversity} and Figure~\ref{fig:diversity_rotation}.
Our method is also realistic in a few ways. Unlike the synthesis or rendering approaches, we construct the whole-body objects using only the real LiDAR observations collected with the real LiDAR system. Our approach can be relatively noisy but the constructed objects are guaranteed to be realistic. Another factor that makes our augmentation realistic is the adoption of HPR. It enables application of occlusion without any concern on the computational burden. In consequence, random placements of whole-body objects can become realistic by applying self-occlusion and external-occlusion with HPR.
Despite the advantages listed above, we were not able to consider the environment dependent distributions of the object location, heading angle, object class, object shape, etc. We have simply employed uniform random strategies, and this remains as a limitation of our work.  

\vspace{0.1cm}
\paragraph{Performance of Car Detection:}
Our method provides an excellent performance for pedestrian and cyclist detection as can be seen in Table~\ref{tab:big_comparisons}. The performance for car detection, however, is relatively less desirable. From the analysis tables in Section~\ref{sec:analysis}, it can be observed that the car detection performance can be easily improved by sacrificing the performance of the other two categories. For instance, the car performance can be improved from 84.74\% to 85.47\% by disabling e-HPR (see the last two rows of Table~\ref{tab:hpr_ablation}).
Similar observations can be made from Table~\ref{tab:construction}, \ref{tab:analysis_placement1}, and \ref{tab:analysis_placement2} where customizing the strategies for car detection can be beneficial for improving car detection.
In fact, we can even achieve 86.43\% of car detection performance by 
applying a simple yet effective modification. In Table 14 of Supplementary~E,
we are showing the results for including three additional car objects into each scene. With this modification, the car performance is improved at the cost of the pedestrian performance, where the mean performance remains almost the same (slightly improved from 78.30\% to 78.34\%) and the computational overhead also remains almost the same (the total points per frame only marginally increases from 17,203 to 17,628).

Despite the obvious solutions for improving the car detection performance, we have focused on improving the mean performance only because our goal of this work is to develop a general augmentation scheme for any class of object. We didn't want to focus on tuning or developing class-dependent techniques. Adjusting the object numbers in each training scene (e.g., three more cars) is a way of controlling the performance trade-off over the three object categories, and thus it is presented only for the purpose of discussion. In this work, we have considered the mean performance as the only genuine metric of performance. 
As for the general augmentation schemes, it certainly remains open to investigate if a general modification, such as manipulating intensity of augmented points, can improve the car detection performance without harming the pedestrian and cyclist detection performance.

\vspace{0.1cm}
\paragraph{Waymo dataset:}
In the stage of iterative construction, we fill up the missing points of an object by adding points of other objects of the same class. It means that the objects of the same class should have similar shapes and characteristics for a successful construction. Unfortunately, the objects of Waymo dataset are labeled in a quite coarse manner. For instance, motorcycles and motorcyclists share the label of vehicle. Because of the need for additional labeling, we did not evaluate Waymo dataset. Nonetheless, \textsc{DR.CPO} should be effective for any dataset with decently refined categorization.

\section{Conclusion}
\label{sec:conclusion}
In this study, we proposed an augmentation method that was developed specifically for LiDAR datasets. Thanks to its diversified and realistic nature, \textsc{DR.CPO} can perform on par with no-augmentation and CDA augmentation using only 4.60\% and 37.60\% of the training data, respectively. \textsc{DR.CPO} is model-agnostic and can enhance the performance for voxel-based, point-based, and voxel+point-based models. \textsc{DR.CPO} also provides a superior performance without incurring an additional burden in computation. Experiment and analysis results show that \textsc{DR.CPO} is an excellent solution for pedestrian and cyclist detection but that there is a room for improvement for car detection.

\section*{Acknowledgement}
This work was supported by a National Research Foundation of Korea (NRF) grant funded by the Korea government (MSIT) (No. NRF-2020R1A2C2007139) and in part
by IITP grant funded by the Korea government (MSIT) [NO.2021-0-01343, Artificial Intelligence Graduate School Program (Seoul National University)].

\bibliography{Shin}

\newpage
\appendix

\twocolumn[{%
 \centering
 \LARGE Supplementary materials for the paper \\ ``Diversified and Realistic 3D Augmentation via \\ Iterative Construction, Random Placement, and HPR Occlusion''\\[1.5em]
}]

\section{Implementation Details}
\label{sec:implementation_details_augmentation}

\subsection{Iterative Construction}
\label{sec:supp_iterative_construction}

\subsubsection{A.1.1 Indexing Candidate Objects (Pre-processing)} 

\begin{figure}[h!]
\centering
\includegraphics[width=0.8\columnwidth]{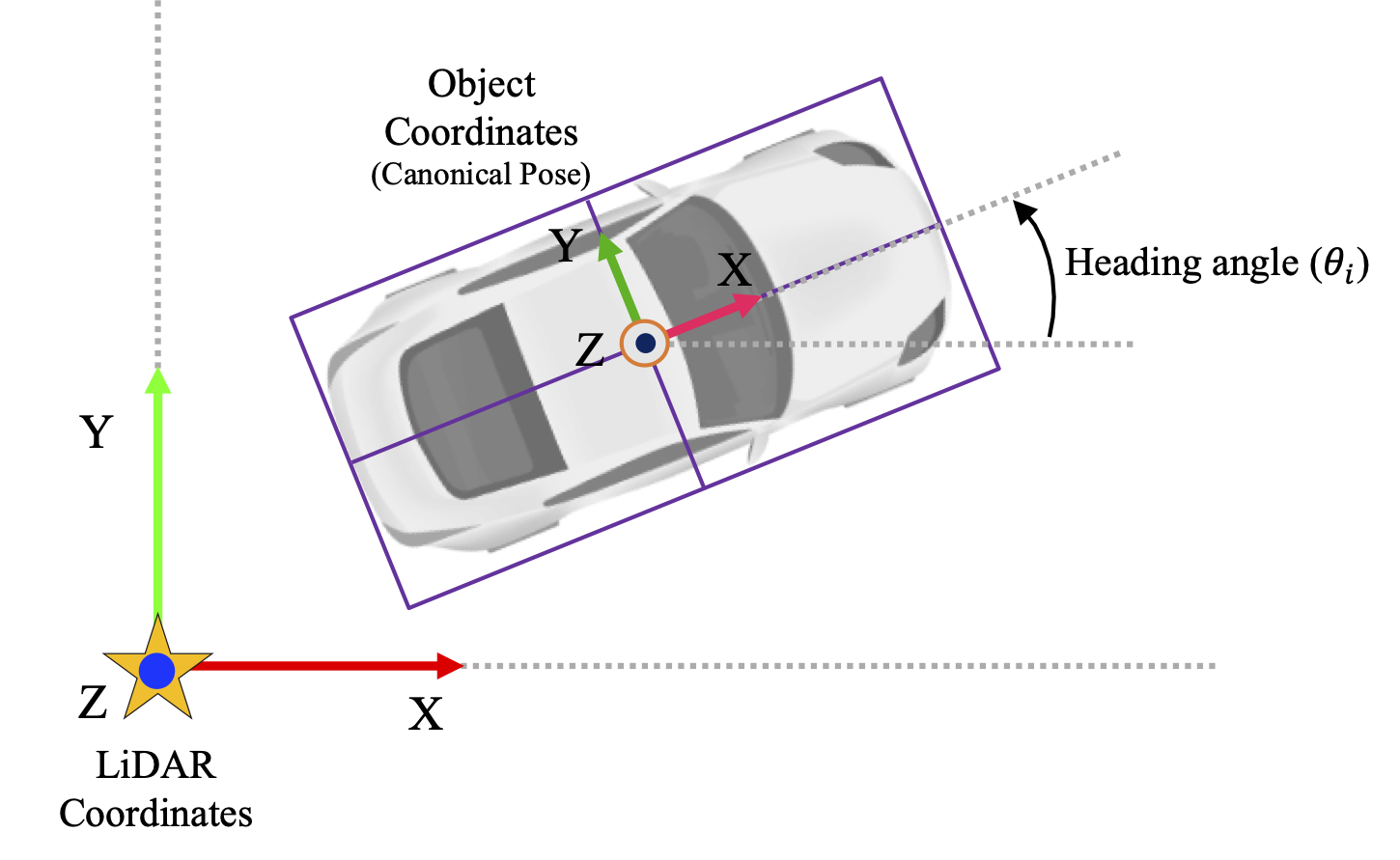}
\caption{\label{fig:Lidar_coordinates.}
An illustration of 3D coordinates. For a raw dataset, we use the LiDAR coordinates. In our study, we apply the object-level operations, including iterative construction and s-HPR, after we convert each object into the canonical pose. In other words, we center the object to have the same heading angle of $\theta_i=0$. 
Thus, when we convert the object into the canonical pose, all objects are toward the same direction.
(The figure is adapted from Figure~\ref{fig:appendix_car} of \cite{geiger2013vision}.)
}
\end{figure}

\paragraph{Canonical Pose: } A whole-body object is constructed in the canonical pose. As shown in Figure~\ref{fig:Lidar_coordinates.}, (1) we first center the object, $(\text{cx}_i, \text{cy}_i, \text{cz}_i)=(0, 0, 0)$, and (2) rotate the object to have the heading angle of $\theta_i=0$.

\paragraph{Size Similarity of Bounding Boxes: }
We calculate the bounding box similarity with $O_i$ for all the objects $O_j$ ($i\neq j$ and $s_i=s_j$) in the ground-truth database to select the best $2K$ candidates. The bounding box similarity is defined as below:
\begin{equation}
    \text{BoxSimilarity}(i,j) = \frac{\text{Vol}(B_i \cap B_j)}{\text{Vol}(B_i \cup B_j)},
\end{equation}
where $B_i$ is the bounding box of $O_i$ in canonical pose and $\text{Vol}(B)$ denotes the volume of $B$.

\paragraph{Partition Density: }
Among the $2K$ candidates, the best $K$ candidates that have high densities for $O_i$'s low density partitions are selected as the final candidates. We define the partitions as a set of non-overlapping subdivided regions of the object $O_i$. A fixed number of partitions is applied for each class. We evaluate the density of an $O_i$'s partition as the number of points in the partition normalized by the maximum number of points observed for the partition over all the same-class objects in the ground-truth database.

\vspace{1cm}

\subsubsection{A.1.2 Constructing a Whole-body Object (Training Time)}
\begin{figure}[h!]
    \centering
    \subfloat[]{\includegraphics[height=0.13\paperheight]{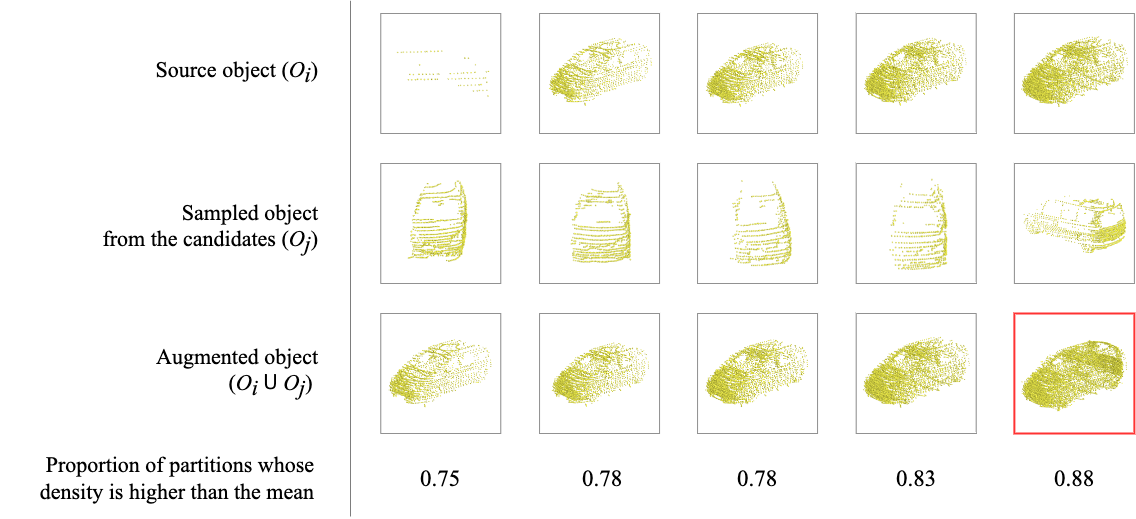}} \\
    \subfloat[]{\includegraphics[height=0.13\paperheight]{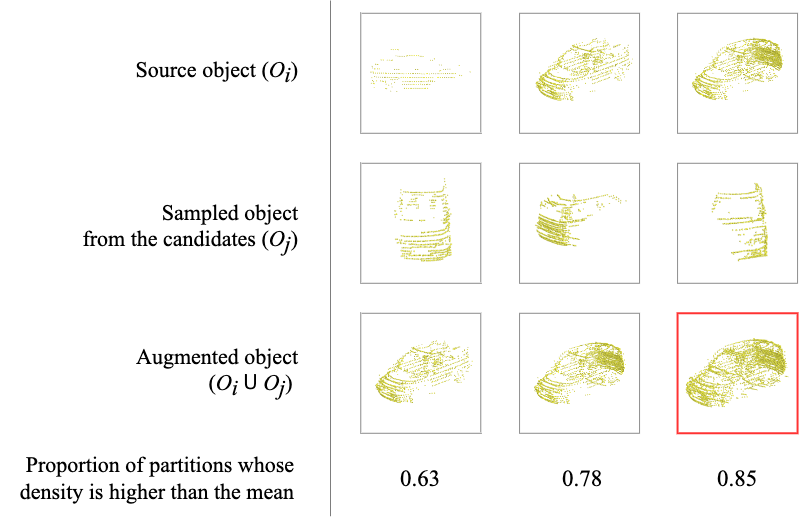}}
    \caption{\label{fig:appendix_car} Two visualization examples of iterative construction for car. Construction ends after 5 iterations in (a), and 3 in (b). On the average, construction ends after 4.38 iterations for the car object.}
\end{figure}

\begin{figure}[h!]
    \centering
    \subfloat[]{\includegraphics[height=0.13\paperheight]{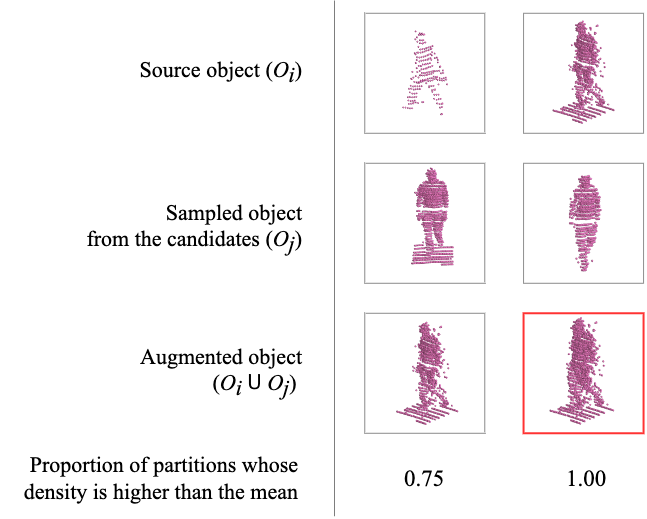}} \\
    \subfloat[]{\includegraphics[height=0.13\paperheight]{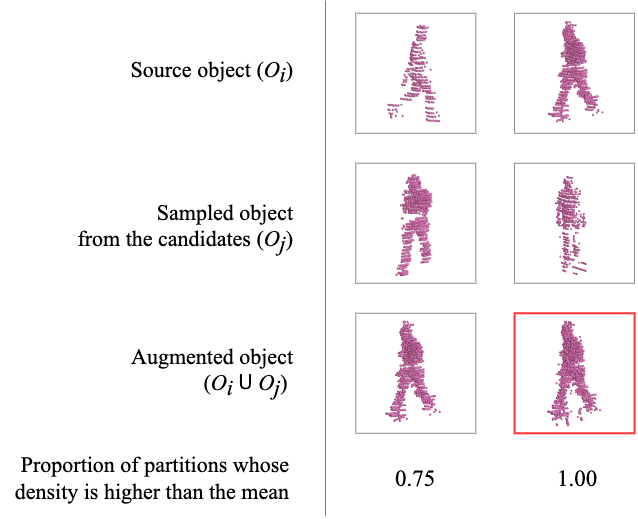}}
    \caption{\label{fig:appendix_ped} Two visualization examples of iterative construction for pedestrian. Construction ends after 2 iterations in both (a) and (b). On the average, construction ends after 2.09 iterations for the pedestrian object.}
\end{figure}

\begin{figure}[h!]
    \centering
    \subfloat[]{\includegraphics[height=0.13\paperheight]{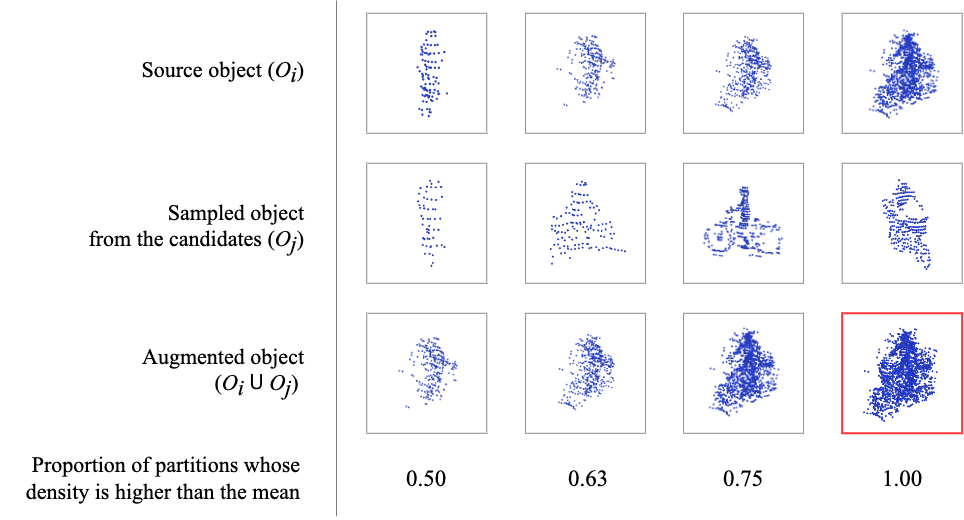}} \\
    \subfloat[]{\includegraphics[height=0.13\paperheight]{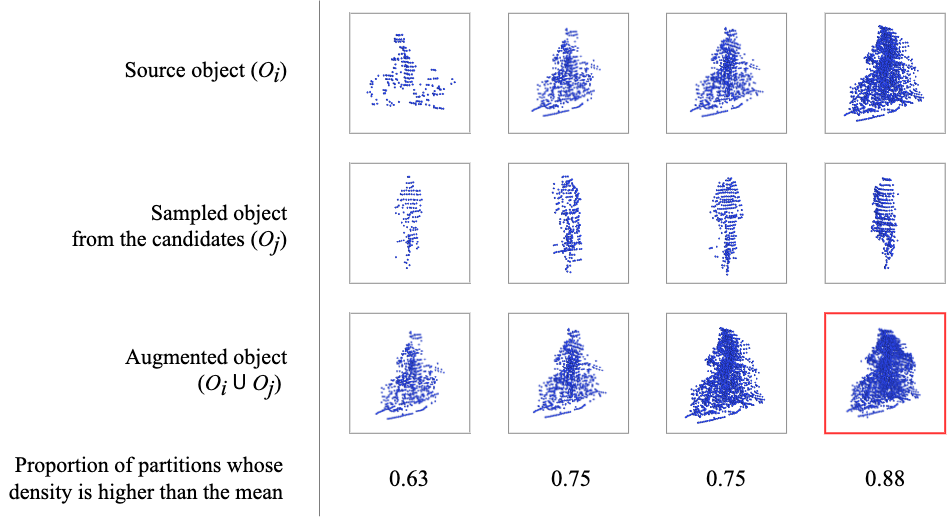}}
    \caption{\label{fig:appendix_cyc} Two visualization examples of iterative construction for cyclist. Construction ends after 4 iterations in both (a) and (b). On the average, construction ends after 3.75 iterations for the cyclist object.}
\end{figure}

\paragraph{\phantom{0}}
As explained in the main part of this work, we iteratively construct a whole-body object until the object $O_i$ can be regarded as a whole-body object. We determine the object to be a whole-body object when the proportion of the high density partitions is at least 85\%, where a partition is considered to be a high density if its density is higher than the mean density of the partition excluding the empty.
We also limit the number of iterations to be at most 20.
We provide examples of iterative construction with KITTI dataset in Figure~\ref{fig:appendix_car}, \ref{fig:appendix_ped}, and \ref{fig:appendix_cyc}. Each column corresponds to a single iteration. For each iteration, we use the augmented object in the previous iteration as the source object. At the bottom, we have also indicated the proportion of partitions whose density is higher than the mean. The final results of the whole-body object construction are indicated with red boxes. For a better illustration, we have chosen a proper heading angle $\theta_i$ for each object. The actual iterative construction process is performed in the canonical pose.

\subsection{Random placement}
The detection range is [0, 70.4]$m$ for the X axis, [$-40$, $40$]$m$ for the Y axis, and [$-3$, $1$]$m$ for the Z axis, and the size for voxelization is [0.05, 0.05, 0.1]$m$ for the three axes. 
In our method, ten additional whole-body objects are added for each of car, pedestrian, and cyclist, and all the objects are placed by applying random location and random rotation within the detection range. By implementing collision avoidance~\cite{yan2018second}, objects with overlapped bounding boxes are excluded.

\subsection{HPR occlusion}
For s-HPR, $R$ value needs to be sufficiently large and we set it as the diagonal length of the bounding box multiplied by 200. After fixing the radius $R$, we use the randomized location to determine the $C$ value. Therefore, $C$ is small for an object located close to the LiDAR and large for an object located far from the LiDAR. For e-HPR, $R$ and $z$ values were chosen dependent on the model. The $x$ and $y$ coordinates were set with respect to the LiDAR origin $(0,0,0)$. The parameters for four main models are listed in Table~\ref{tab:hpr_parameter}.
\begin{table}[t!]
\caption{\label{tab:hpr_parameter} The $R$ and $z$ values for the HPR occlusion.}
\vspace{-0.15cm}
\centering
\footnotesize{
\begin{tabular}{@{}cccccccc@{}}
\toprule
    Model      & R   & z \\ \midrule
    SECOND     & 300,000 & 2.0 \\
    PointRCNN  & 300,000 & 1.0 \\
    PV-RCNN    & 100,000 & 0.0 \\
    PV-RCNN++  & 100,000 & 0.0 \\
    \bottomrule
\end{tabular}

\vspace{-0.25cm}
}
\end{table}
\begin{figure}[t!]
     \centering
     \includegraphics[width=0.26\textwidth]{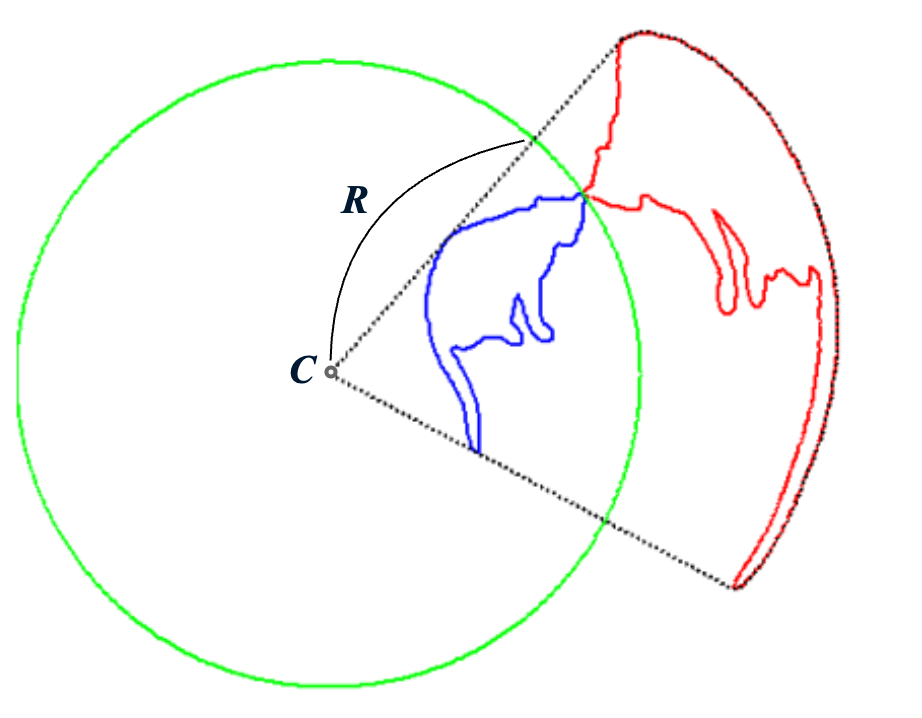}
     \caption{(Adapted from Figure~\ref{fig:sparsity_with_distance} of \cite{katz2007direct}) HPR can determine the points that are visible from the viewpoint $C$. In the figure, the points inside the sphere of radius $R$ correspond to the original object and the points outside the sphere correspond to the spherically flipped object. HPR determines an original point to be visible from $C$ if its spherically flipped point lies on a convex hull.}
     \label{fig:hpr}
\end{figure}

\subsection{Conventional Data Augmentation}
For the Global Data Augmentation~(GDA), we utilized random flipping along the X axis, global scaling with a random scaling factor sampled from [0.95, 1.05], global rotation around the Z axis with a random angle sampled from [$-\pi$/4,$\pi$/4]. We also conduct the Ground-Truth Sampling~(GTS) to randomly copy and paste objects from the ground-truth database. For each of car, pedestrian, and cyclist, we copy-and-paste 20, 15, and 15 objects, respectively, for each training frame. 

\section{Experimental Details}
\label{sec:implementation_details_experiment}
\subsection{Model and Training Details: }
For the implementation of the three models~(SECOND, PointRCNN, and PV-RCNN) and Conventional Data Augmentation~(CDA), we followed the open source code base https://github.com/open-mmlab/OpenPCDet~\cite{openpcdet2020} and the original hyper-parameter configurations in there.
For PV-RCNN+, however, configuration of KITTI is not available. 
Therefore, we slightly modified PV-RCNN++'s Waymo configuration file by 
comparing it with the PV-RCNN's KITTI configuration file. All the modifications that we have made can be listed as the following. 
\begin{itemize}
\item WEIGHT\_DECAY is set to 0.01 instead of 0.001.
\item NUM\_REDUCED\_CHANNELS in raw\_points is set to 1 instead of 2, because KITTI has only one intensity channel in contrast to Waymo with two intensity channels.
\item NUM\_KEYPOINTS is set to 2048 instead of 4096, because average points per frame in KITTI is less than half of those in Waymo.
\end{itemize}

\subsection{Dataset and Evaluation Metric: } 
The models are evaluated using the validation dataset with the standard metric - the average precision (AP) with an IOU threshold value for the car, pedestrian, and cyclist classes of 0.7, 0.5, and 0.5, respectively. As is done most of the other works, we calculated the AP performances with 40 recall positions (R40) on three difficulties (easy, moderate, and hard) that are typically determined based on the object size, occlusion, and truncation levels.

\clearpage
\onecolumn
\section{Additional KITTI Examples}
\label{sec:appendix_examples}
We provide KITTI examples of the random placement, s-HPR, and e-HPR.

\begin{figure*}[h!]
    \label{fig:appendix_ehpr}
    \centering
    \subfloat[]{\includegraphics[width=0.9\textwidth]{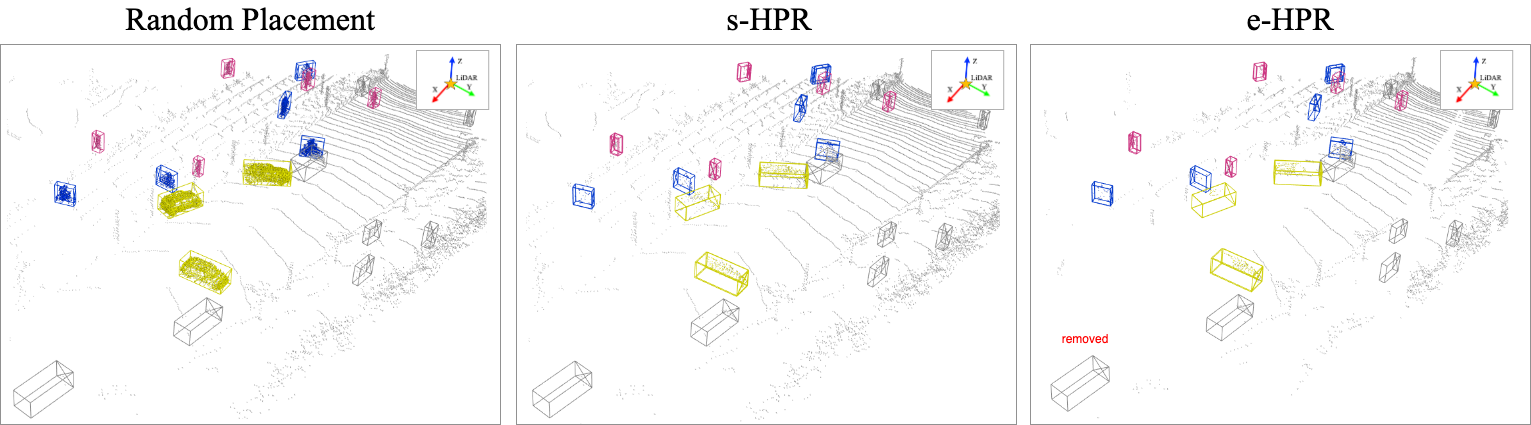}}\\
    \subfloat[]{\includegraphics[width=0.9\textwidth]{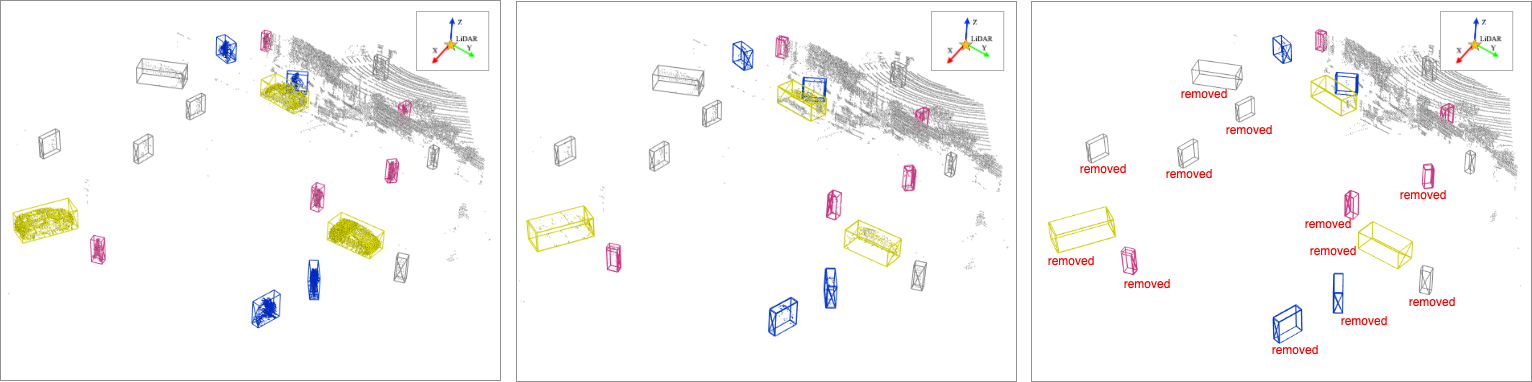}}\\
    \subfloat[]{\includegraphics[width=0.9\textwidth]{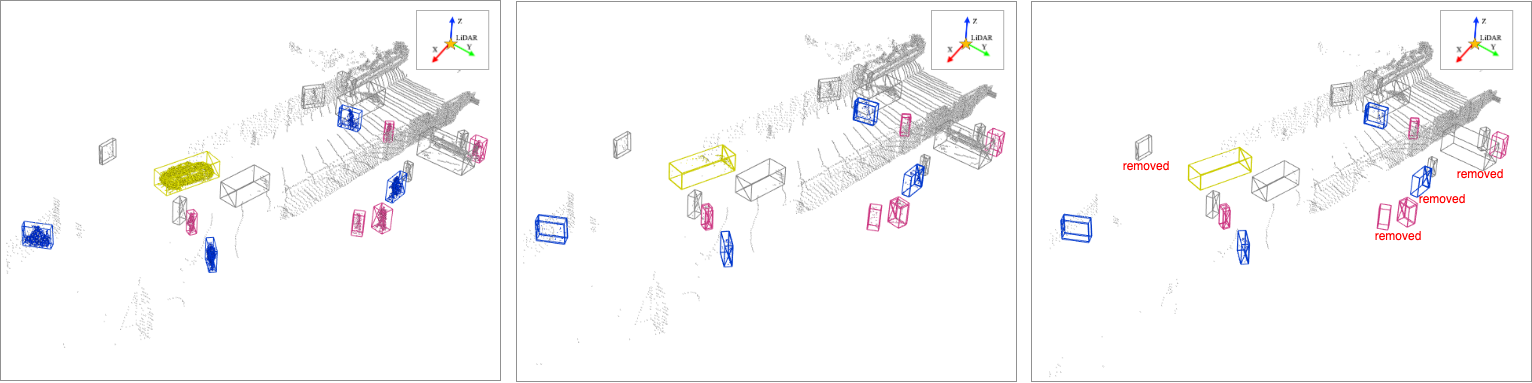}}\\
    \subfloat[]{\includegraphics[width=0.9\textwidth]{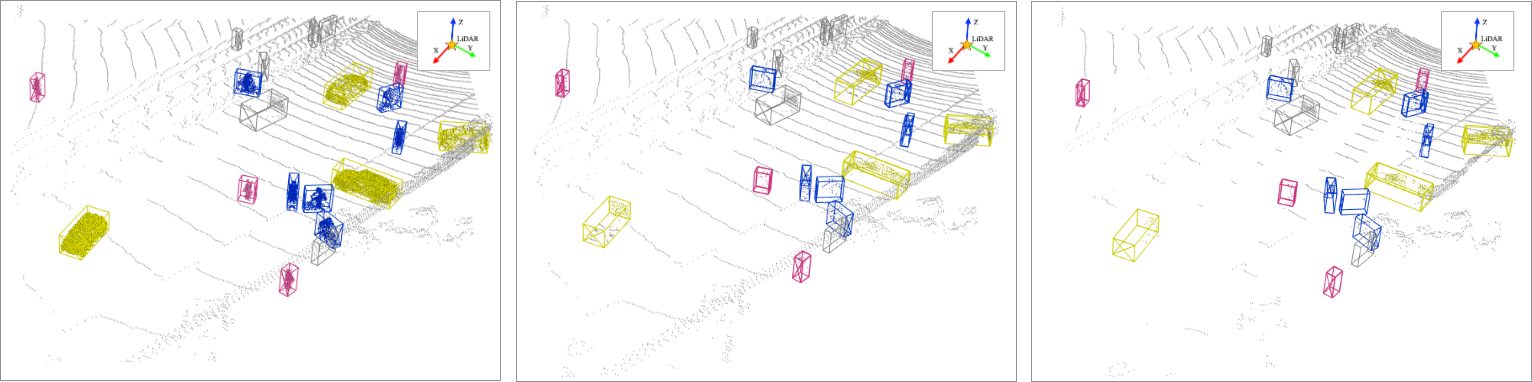}}
    \caption{Examples of random placement (Left), s-HPR (Middle), and e-HPR (Right) for KITTI dataset.}
\end{figure*}

\clearpage
\section{Full Results}
\label{sec:additional_results}

\begin{table*}[h!]
\centering
\caption{Full results of Table~\ref{tab:model_agnostic}.}
\resizebox{0.9\textwidth}{!}{
\begin{tabular}{@{}llccccccccccccc@{}}
\toprule
\multicolumn{2}{c}{\multirow{2}{*}{Method}}                               & \multicolumn{3}{c}{Car} & \multicolumn{3}{c}{Pedestrian} & \multicolumn{3}{c}{Cyclist} & \multicolumn{3}{c}{Overall} & \multirow{2}{*}{Mean}           \\\cmidrule(lr){3-5}\cmidrule(lr){6-8}\cmidrule(lr){9-11}\cmidrule(lr){12-14}
\multicolumn{1}{c}{} & \multicolumn{1}{c}{}              & E        & M        & D        & E           & M          & D          & E          & M         & D         & Car     & Ped     & Cyc     &                \\ \midrule
Voxel-based          & SECOND~\cite{xu2022behind}      & \textbf{90.97}    & 79.94    & 77.09    & \textbf{58.01}       & 51.88      & 47.05      & 78.50      & 56.74     & 52.83     & 82.67   & 52.31   & 62.69   & 65.89          \\
                     & SECOND with \textsc{DR.CPO}                     & 90.30    & \textbf{81.10}    & \textbf{78.04}    & 57.35       & 51.02      & 46.65      & \textbf{87.48}      & \textbf{68.94}     & \textbf{64.84}     & \textbf{83.15}   & 51.67   & \textbf{73.75}   & \textbf{69.53} \\ \midrule
Point-based          & Point-RCNN~\cite{hu2021pattern} & 90.10    & 80.41    & \textbf{78.00}    & 64.18       & 56.71      & 49.86      & \textbf{91.72}      & 72.47     & 68.18     & 82.84   & 56.92   & 77.46   & 72.40          \\
                     & Point-RCNN with \textsc{DR.CPO}                 & 89.51    & 79.82    & 77.75    & 67.05       & \textbf{59.86}      & \textbf{52.66}      & 91.48      & \textbf{73.42}     & \textbf{69.74}     & 82.36   & \textbf{59.86}   & \textbf{78.22}   & \textbf{73.48} \\ \midrule

Voxel+point-based   & PV-RCNN~\cite{xu2022behind}        & \textbf{92.57}  & \textbf{84.83} & \textbf{82.69} & 64.26         & 56.67          & 51.91                                         & 88.88          & 71.95          & 66.78          & \textbf{86.70} & 57.61 & 75.87 & 73.39 \\ 
                    & PV-RCNN with \textsc{DR.CPO}                    & 91.07           & 81.50          & 81.68          & \textbf{68.66} & \textbf{61.20} & \textbf{57.11} & \textbf{93.77} & \textbf{77.06} & \textbf{73.57} & 84.75 & \textbf{62.33} & \textbf{81.47} & \textbf{76.18} \\  \bottomrule
\end{tabular}
}
\vspace{-0.4cm}
\end{table*}

\begin{table*}[h!]
\caption{Full results of Table~\ref{tab:construction}.}
\centering
\resizebox{0.9\textwidth}{!}{
\begin{tabular}{@{}ccccccccccccccc@{}}
\toprule
\multirow{2}{*}{Mirroring} & \multirow{2}{*}{Adding Candidates} & \multicolumn{3}{c}{Car} & \multicolumn{3}{c}{Pedestrian} & \multicolumn{3}{c}{Cyclist} & \multicolumn{3}{c}{Overall} & \multirow{2}{*}{Mean}\\ \cmidrule(lr){3-5}\cmidrule(lr){6-8}\cmidrule(lr){9-11}\cmidrule(lr){12-14}
 & & E      & M      & D     & E        & M        & D        & E       & M       & D       & Car     & Ped     & Cyc     &                    \\ \midrule
No & None     & 91.45 & 82.13 & 81.66 & 67.77 & 62.94 & 58.32 & 92.18 & 75.02 & 72.06 & 85.08 & 63.01 & 79.75 & 75.95              \\ 
Yes & None    & 91.88 & 82.10 & 81.58 & 68.59 & 62.39 & 58.94 & 92.37 & 75.15 & 71.56 & 85.18 & 63.31 & 79.69 & 76.06         \\
Yes & Single & 91.68  & 82.18  & 81.91 & 73.21    & 67.56    & 62.78    & 93.34   & 75.67   & 72.69   & 85.25   & 67.85   & 80.57   & 77.89         \\
Yes & Multiple & 91.27  & 81.67  & 81.28 & 73.23    & 67.66    & 62.96    & 93.08   & 78.21   & 75.32   & 84.74   & 67.95   & 82.20   & 78.30         \\ \bottomrule
\end{tabular}
}
\vspace{-0.4cm}
\end{table*}

\begin{table}[h!]
\caption{Full results of Table~\ref{tab:analysis_placement1}.}
\centering
\resizebox{0.9\columnwidth}{!}{
\begin{tabular}{@{}ccccccccccccccccc@{}}
\toprule
    \multirow{2}{*}{\begin{tabular}[c]{@{}l@{}}Random \\ Location\end{tabular}}  & \multirow{2}{*}{\begin{tabular}[c]{@{}l@{}}Random \\ Rotation\end{tabular}}   & \multicolumn{3}{c}{Car} & \multicolumn{3}{c}{Pedestrian}      & \multicolumn{3}{c}{Cyclist}   & \multicolumn{3}{c}{Overall} & \multirow{2}{*}{Mean} \\\cmidrule(lr){3-5}\cmidrule(lr){6-8}\cmidrule(lr){9-11}\cmidrule(lr){12-14}
     & & E & M & D & E & M & D & E & M & D & E & M & D & \\\midrule
    
                    &              & 91.46 & 81.50 & 79.98 & 70.63 & 65.02 & 61.10 & 91.32 & 75.98 & 72.25    & 84.31 &  65.58        & 79.85     & 76.58 \\
     \checkmark     &                   & 91.37 & 81.51 & 79.95 & 72.36 & 66.65 & 62.03 & 90.44 & 75.20 & 72.31 & 84.28 &  67.01        & 79.32     & 76.87 \\ 
                    & \checkmark        & 91.91 & 82.21 & 81.85 & 71.06 & 65.81 & 61.12 & 93.69 & 79.28 & 76.52 & 85.33 &  66.00        & 83.16     & 78.16 \\
     \checkmark     & \checkmark        & 91.27 & 81.67 & 81.28 & 73.23 & 67.66 & 62.96 & 93.08 & 78.21 & 75.32 & 84.74 &  67.95        & 82.20     & 78.30 \\
    \bottomrule
\end{tabular}
}
\vspace{-0.4cm}
\end{table}

\begin{table*}[h!]
\caption{Full results of Table~\ref{tab:analysis_placement2}.}
\centering
\resizebox{0.9\textwidth}{!}{
\begin{tabular}{@{}cccccccccccccc@{}}
\toprule
\multirow{2}{*}{Rotation range} & \multicolumn{3}{c}{Car} & \multicolumn{3}{c}{Pedestrian} & \multicolumn{3}{c}{Cyclist} & \multicolumn{3}{c}{Overall} & \multirow{2}{*}{Mean}        \\ \cmidrule(lr){2-4}\cmidrule(lr){5-7}\cmidrule(lr){8-10}\cmidrule(lr){11-13}
                & E      & M      & D     & E        & M        & D        & E       & M       & D       & Car     & Ped     & Cyc     &  \\ \midrule
$(-0.25\pi, +0.25\pi)$       & 91.42  & 82.26  & 81.70 & 70.60    & 63.73    & 58.92    & 93.54   & 75.87   & 73.21   & 85.13   & 64.41   & 80.87   & 76.81   \\
$(-0.50\pi, +0.50\pi)$       & 91.80  & 82.32  & 81.96 & 71.10    & 64.82    & 59.04    & 92.47   & 75.74   & 73.19   & 85.36   & 64.99   & 80.46   & 76.94   \\
$(-0.75\pi, +0.75\pi)$     & 91.34  & 82.25  & 80.08 & 71.37    & 65.33    & 60.30    & 94.57   & 77.08   & 73.18   & 84.56   & 65.67   & 81.61   & 77.28   \\
$(-1.00\pi, +1.00\pi)$     & 91.27  & 81.67  & 81.28 & 73.23    & 67.66    & 62.96    & 93.08   & 78.21   & 75.32   & 84.74   & 67.95   & 82.20   & 78.30   \\ \bottomrule
\end{tabular}
}
\vspace{-0.4cm}
\end{table*}

\begin{table}[thb!]
\caption{Full results of Table~\ref{tab:hpr_ablation}}
\centering
\resizebox{0.9\columnwidth}{!}{
\begin{tabular}{@{}ccccccccccccccccc@{}}
\toprule
    \multirow{2}{*}{s-HPR} & \multirow{2}{*}{e-HPR} & \multicolumn{3}{c}{Car} & \multicolumn{3}{c}{Pedestrian} & \multicolumn{3}{c}{Cyclist} & \multicolumn{3}{c}{Overall} & \multirow{2}{*}{Mean} \\ \cmidrule(lr){3-5}\cmidrule(lr){6-8}\cmidrule{9-11}\cmidrule(lr){12-14}
      & & E              & M              & D              & E              & M              & D              & E              & M              & D              & Car            & Ped            & Cyc            &         \\ \midrule
                &               & 91.79 & 82.13 & 80.01 & 57.66 & 52.18 & 47.16 & 93.25 & 73.72 & 69.84 & 84.64 & 52.33 & 78.94 & 71.97 \\
                & \checkmark    & 91.50 & 81.82 & 81.61 & 69.75 & 64.58 & 60.17 & 90.16 & 76.74 & 73.77 & 84.98 & 64.83 & 80.22 & 76.68 \\
     \checkmark &               & 91.72 & 82.70 & 82.00 & 69.39 & 63.72 & 57.63 & 95.95 & 78.97 & 74.24 & 85.47 & 63.58 & 83.05 & 77.37 \\
     \checkmark & \checkmark    & 91.27 & 81.67 & 81.28 & 73.23 & 67.66 & 62.96 & 93.08 & 78.21 & 75.32 & 84.74 & 67.95 & 82.20 & 78.30 \\ 
    \bottomrule
\end{tabular}
}
\vspace{-0.4cm}
\end{table}

\section{Additional Results}
\label{sec:Supplementary_E}

\begin{table}[h!]
\caption{\label{tab:additional_experiment}Performance trade-off: by adding three additional cars into each training scene, the car detection performance can be improved at the cost of the pedestrian detection performance. Note that the mean performance remains almost the same.
}

\centering
\resizebox{0.9\columnwidth}{!}{
\begin{tabular}{@{}lccccccccccccc@{}}
\toprule
\multirow{2}{*}{Method} & \multicolumn{3}{c}{Car} & \multicolumn{3}{c}{Pedestrian} & \multicolumn{3}{c}{Cyclist} & \multicolumn{3}{c}{Overall} & \multirow{2}{*}{Mean}        \\ \cmidrule(lr){2-4}\cmidrule(lr){5-7}\cmidrule(lr){8-10}\cmidrule(lr){11-13}
                & E      & M      & D     & E        & M        & D        & E       & M       & D       & Car     & Ped     & Cyc     &  \\ \midrule
PVRCNN++ baseline          & 91.72 & 84.82 & 82.03 & 66.00 & 60.05 & 54.70 & 91.39 & 70.14 & 66.92 & 86.19 & 60.25 & 76.15 & 74.20 \\
DR.CPO                     & 91.27 & 81.67 & 81.28 & 73.23 & 67.66 & 62.96 & 93.08 & 78.21 & 75.32 & 84.74 & \textbf{67.95} & 82.20 & 78.30 \\
DR.CPO (3 additional cars) & 92.14	& 84.54	& 82.60	& 71.67	& 65.63	& 60.78	& 93.88	& 78.35 & 75.44 & \textbf{86.43} & 66.03 & 82.56 & 78.34 \\ \bottomrule
\end{tabular}
}
\vspace{-0.4cm}
\end{table}

\twocolumn

\end{document}